\documentclass{article}

\usepackage[numbers,compress]{natbib}

\usepackage[main, final]{neurips_2026}

\usepackage[utf8]{inputenc}
\usepackage[T1]{fontenc}
\usepackage{hyperref}
\usepackage{url}
\usepackage{booktabs}
\usepackage{amsfonts}
\usepackage{nicefrac}
\usepackage{microtype}
\usepackage[table]{xcolor}
\usepackage{xspace}

\usepackage[dvipsnames,table]{xcolor}
\definecolor{ourblue}{rgb}{0.368,0.507,0.71}
\definecolor{ourred}{rgb}{0.923,0.386,0.209}
\definecolor{url}{HTML}{d95225}
\hypersetup{
    final,
    colorlinks,
    linkcolor=ourred,
    citecolor=ourblue,
    urlcolor=url,
}

\usepackage{tocloft}
\usepackage{amsmath}
\usepackage{amssymb}
\usepackage{mathtools}
\usepackage{amsthm}
\newtheorem{definition}{Definition}
\usepackage{dsfont}
\usepackage{enumitem}
\usepackage{tikz,ifthen}

\usepackage{graphicx}
\usepackage{dcolumn}
\usepackage{multirow}
\usepackage{float}
\usepackage{caption}
\usepackage{subcaption}
\usepackage{bm}
\usepackage{wrapfig}

\newcommand{\method}{\textsc{Loot}\xspace}
\newcommand{\cev}[1]{\reflectbox{$\vec{\reflectbox{$#1$}}$}}
\title{Empowering Masked Diffusion Models to Self-Correct with Leave-One-Out Transformers}

\author{
  Sofian Zalouk
  \And Vincent Counathe
  \And  Paul Jünger
  \And  Daniel Cao
  \And  Daniel Zlotnick\vspace{.3em}
  \and \textbf{Kilian Q. Weinberger}
  \hspace{.4em} \textbf{Christopher De Sa}\vspace{.4em}\\
  Department of Computer Science \\
  Cornell University\vspace{.4em} \\
  \texttt{ \{saz43, vc383, pj287, dyc33, dz344, kqw4, cmd353\}@cornell.edu}
}

\begin{document}

\maketitle

\begin{abstract}
Masked diffusion models (MDMs) have emerged as a promising paradigm for language modeling, enabling parallel token generation with competitive performance. Yet standard MDMs suffer from two fundamental limitations: (1) inability to revise tokens, and (2) sparse training signal. We revisit hollow transformers, which address both limitations by predicting each token from its surrounding context, independently of the input at that position. We introduce the Leave-One-Out Transformer (\textsc{LOOT}), a more efficient and expressive architecture that advances hollow transformers through two complementary advances. First, we redesign the merging mechanism to enable frequent interaction between left and right contexts, substantially improving predictive performance over prior hollow transformers. Second, we derive a lower-variance estimator of the MDM NELBO that supervises all tokens and improves validation perplexity. Together, these advances yield a competitive standalone diffusion language model capable of effective self-correction during inference. On LM1B, \textsc{LOOT} matches the strongest dense-supervision MDM baseline in validation perplexity. On OpenWebText, \textsc{LOOT} achieves higher MAUVE and better generation quality--diversity tradeoffs than prior remasking methods, using $8\times$ fewer forward evaluations and $3.5\times$ fewer inference FLOPs. Code and checkpoints are available at \url{https://github.com/szalouk/loot-diffusion}.
\end{abstract}

\section{Introduction}

Diffusion language models (DLMs) offer a fundamentally different approach to text generation from autoregressive (AR) models. Rather than generating tokens sequentially from left to right, DLMs iteratively denoise entire sequences, allowing multiple tokens to be predicted in parallel. Masked diffusion models (MDMs) have emerged as a leading approach to diffusion for language, with recent models scaling up to 100B parameters \citep{bie2025llada20scalingdiffusionlanguage}. Yet standard MDMs lose a defining appeal of diffusion: the ability to iteratively refine earlier predictions. MDMs generate text by progressively replacing \texttt{[MASK]} tokens with model predictions. Once a token is unmasked, it remains fixed, so errors cannot be corrected and propagate through the rest of the generation process. The standard MDM training objective is similarly restrictive. The training loss is computed only on masked positions, leaving visible tokens without direct supervision. Thus, standard MDMs face two limitations we aim to address: (1) \textit{inability to self-correct} during inference, and (2) \textit{sparse training signal}.

\begin{figure*}[t]
    \centering
    \hspace{-1.7cm}
    \includegraphics[width=0.8\textwidth]{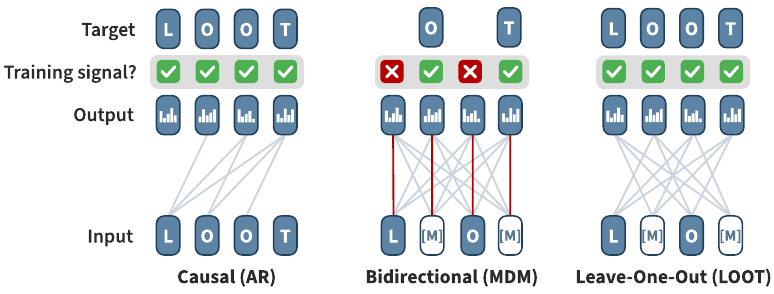}
    \caption{Dependencies of output tokens on input tokens in causal, bidirectional, and leave-one-out models. Unlike standard bidirectional MDMs, the output of \method{} at index $i$ does not depend on the input at index $i$, enabling loss optimization even on unmasked tokens.}
    \vspace{-1em}
    \label{fig:dependency_patterns}
\end{figure*}

Addressing both limitations is challenging because standard MDMs use bidirectional transformers, allowing each prediction to depend on all input tokens. Consequently, we cannot extend the training loss to visible tokens, since the model could simply copy the correct answer rather than learning to predict it from context. Instead, we overcome this obstacle by modeling the conditional distribution $p_\theta(x_i \mid \mathbf{x}_{-i})$ at every position $i$, ensuring that each prediction is independent of its corresponding input token. This \emph{leave-one-out property} (Figure~\ref{fig:dependency_patterns}) has two key consequences: (1) every token, masked or visible, becomes a genuine prediction task over possible replacements that can be supervised during training, and (2) previously generated tokens can be revised during inference. Hollow transformers~\citep{sun2022score,zhao2024informed} enforce this property by design, but have struggled to achieve competitive performance as standalone diffusion language models. We revisit hollow transformers and introduce two complementary advances in architecture and training to bridge this performance gap.

To this end, we introduce the \textbf{Leave-One-Out Transformer (\method{})}, a more expressive and computationally efficient hollow architecture. First, we redesign the merging mechanism of hollow transformers to enable more frequent interaction between left and right contexts (Section~\ref{sec:method-architecture}). Our ablations show that more frequent merging substantially improves language modeling performance, suggesting that repeated interaction allows the model to progressively refine its representations of the surrounding context. \method{} makes frequent merging efficient by avoiding the expanded hidden representations used in prior architectures, enabling full parameter sharing and sliding-window attention compatible with standard FlashAttention kernels. At the largest model size we evaluate, this allows \method{} to merge left and right contexts at every layer, whereas HollowDiff~\citep{zhao2024informed} can afford only two merges at comparable FLOPs. Second, we exploit the leave-one-out property to derive a new, lower-variance, all-position estimator of the MDM NELBO that improves the training of hollow transformers (Section~\ref{sec:method-objective}).

Together, these advances yield better validation perplexity than prior hollow transformers at matched parameter counts and FLOPs, while establishing \method{} as a competitive standalone diffusion language model capable of effective self-correction. On OpenWebText, \method{} surpasses standard MDMs and ReMDM~\citep{wang2025remasking} in generation quality, as measured by generative-perplexity--entropy frontiers and MAUVE, while requiring $8\times$ fewer sampling steps than ReMDM.

\vspace{-1em}
\section{Background}
\label{sec:background}

Diffusion models learn to reverse a pre-defined noising process that gradually transforms data into noise. D3PM extended diffusion to discrete data~\citep{austin2023structureddenoisingdiffusionmodels}, and later work extended discrete diffusion to continuous time~\citep{campbell2022continuoustimeframeworkdiscrete, lou2024sedd, sahoo2024mdlm}. Two common choices of noising process are \emph{absorbing-state} diffusion, which replaces tokens with a special \texttt{[MASK]} token, and \emph{uniform} diffusion, which corrupts tokens by replacing them with random tokens from the vocabulary. Uniform diffusion allows previously generated tokens to be revised, but absorbing-state diffusion has proved particularly effective for language modeling, owing in part to its simple training objective and empirical success~\citep{sahoo2024mdlm, ye2025dream7bdiffusionlarge, nie2025largelanguagediffusionmodels}.

\subsection{Masked Diffusion Models}
\label{sec:bg-mdm}

Consider a sequence $\mathbf{x}=(x_1,\ldots,x_N)$ of $N$ tokens drawn from a vocabulary $\mathcal{V}$. Masked diffusion defines a forward noising process that independently replaces each token with \texttt{[MASK]} with probability $1-\alpha_t$, where $t\in[0,1]$ and $\alpha_t$ is a monotonically decreasing noise schedule satisfying $\alpha_0=1$ and $\alpha_1=0$. Throughout this paper, we use $M_t\subseteq\{1,\ldots,N\}$ to denote the set of masked positions at time $t$, and $\mathbf{x}[M_t]$ to denote the resulting partially masked sequence. The forward process thus interpolates between clean data at $t=0$ and a fully masked sequence at $t=1$. Crucially, once a token is masked in the forward process, it remains masked.

\paragraph{Training.}
The MDM training objective is a continuous-time NELBO that reduces to a weighted cross-entropy over masked positions~ \citep{sahoo2024mdlm, shi2025simplifiedgeneralizedmaskeddiffusion}:
\begin{equation}
    \mathcal{L}_{\text{MDM}}(\theta)
    = \mathbb{E}_{t,\mathbf{x},M_t}
    \left[
        \frac{\alpha_t'}{1-\alpha_t}
        \sum_{i\in M_t}
        \log p_{\theta}\!\left(x_i\mid\mathbf{x}[M_t], t\right)
    \right],
    \label{eq:mdm-loss}
\end{equation}
where $t\sim\mathcal{U}[0,1]$ and $\alpha_t'=d\alpha_t/dt$. Since the optimal $p_{\theta}$ is independent of $t$ given the masked sequence~\citep{zheng2025maskeddiffusionmodelssecretly, ou2025your, kim2025trainworstplanbest}, we follow standard practice and drop explicit time conditioning, writing the denoiser as $p_\theta(x_i\mid\mathbf{x}[M_t])$. Importantly, the MDM loss in Eq.~\eqref{eq:mdm-loss} is evaluated only at masked positions, leaving visible tokens without direct supervision.

\paragraph{Sampling.}
Generation starts from a fully masked sequence and progressively reveals tokens. Given a partially masked sequence $\mathbf{z}_t$, MDMs predict a distribution over tokens at each masked position. In practice, multiple positions can be unmasked in parallel through $\tau$-leaping~\citep{lou2024sedd, campbell2022continuoustimeframeworkdiscrete}, allowing generation in fewer steps than the sequence length. However, tokens that are unmasked in the same step are sampled independently conditioned on $\mathbf{z}_t$, and may be mutually inconsistent~\citep{liu2024discrete}. Previously unmasked tokens cannot be revised, so errors may propagate through subsequent generation steps.

\subsection{Hollow Transformers}
\label{sec:bg-hollow}

Hollow transformers~\citep{sun2022score, zhao2024informed, ou2026discrete} offer an architectural means of addressing both limitations of MDMs. They satisfy the \emph{leave-one-out property} by design: the prediction at position $i$ depends only on $\mathbf{x}_{-i}$ and not on the input token $x_i$. A hollow transformer therefore computes the conditional distributions $p_\theta(x_i \mid \mathbf{x}_{-i})$ at every position in a single forward pass. As a result, both masked and visible tokens can be supervised during training, and previously generated tokens can be revised during inference.

To achieve this property, \citet{sun2022score} use two causal streams: a \emph{forward stream} $\vec{\mathbf{f}}$ whose representation at position $i$ depends only on earlier tokens $j \le i$, and a \emph{backward stream} $\cev{\mathbf{b}}$ whose representation at position $i$ depends only on later tokens $k \ge i$. To recover a bidirectional representation at position $i$ that satisfies the leave-one-out property, hollow transformers merge the representations of left context $\vec{\mathbf{f}}_{<i}$ and right context $\cev{\mathbf{b}}_{>i}$. \citet{sun2022score} only merge at the final layer, while HollowDiff~\citep{zhao2024informed} introduces a third \emph{output stream} that merges at intermediate layers.

Despite these advances, hollow transformers have struggled to match the performance of bidirectional transformers. For one, HollowDiff reports weaker performance than a standard MDM and achieves its best results by deploying the hollow model as a corrector for a separate MDM predictor model. This predictor-corrector setup incurs additional training and inference costs. We aim to bridge this gap through two complementary improvements. First, we redesign the merging mechanism to enable efficient, frequent interaction between left and right contexts. Our ablations show that merging at every layer substantially improves language modeling performance, suggesting that repeated interaction allows the model to progressively refine its representations of the surrounding context. Second, we derive a lower-variance estimator of the MDM NELBO that improves training efficiency.

\section{Leave-One-Out Language Modeling}
\label{sec:method}

\begin{figure*}[t]
    \centering
    \includegraphics[width=\textwidth]{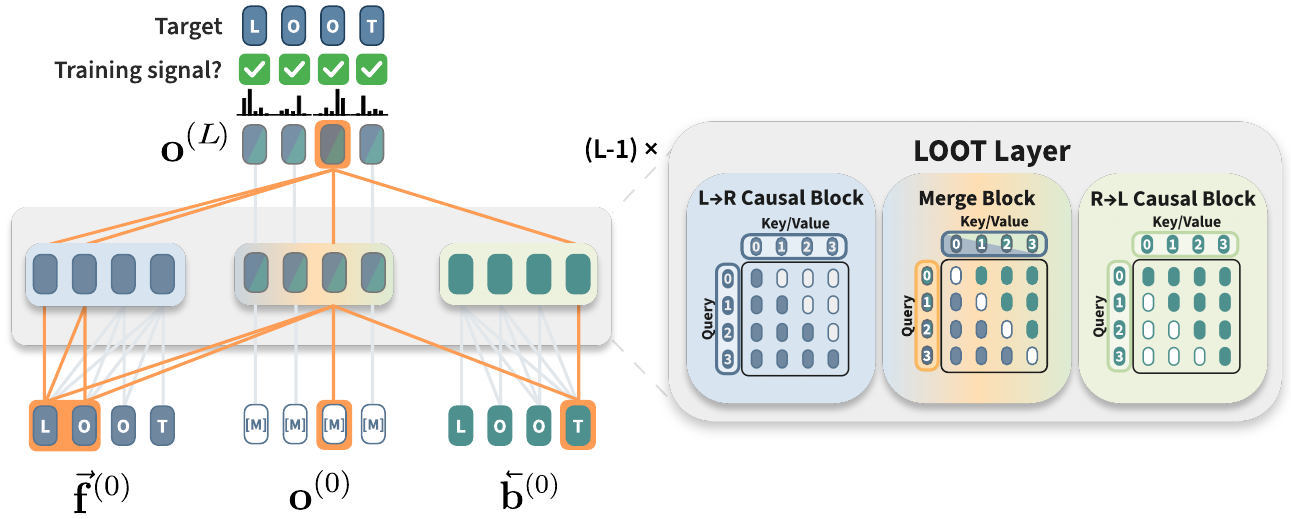}
    \caption{\textbf{Overview of the Leave-One-Out Transformer (\method{}).} Our goal is to predict every token $x_i$ from its surrounding context $\mathbf{x}_{-i}$, independently of the input token at position $i$. \method{} maintains three streams: a left-to-right stream $\vec{\mathbf{f}}$, a right-to-left stream $\cev{\mathbf{b}}$, and an output stream $\mathbf{o}$ that merges the left context $\vec{\mathbf{f}}_{<i}$ and right context $\cev{\mathbf{b}}_{>i}$ at every layer. A stack of $(L{-}1)$ \method{} layers followed by a final merge produces the conditionals $p_\theta(x_i \mid \mathbf{x}_{-i})$ at every position in a single forward pass. This enables supervision at every token during training and self-correction during inference.}
    \label{fig:loo_method}
\end{figure*}

We introduce the Leave-One-Out Transformer (\method{}), a hollow architecture that computes the conditional distributions $p_\theta(x_i \mid \mathbf{x}_{-i})$ at every position $i$ in a single forward pass (Figure~\ref{fig:loo_method}). By predicting each token $x_i$ independently of the input at that position, \method{} enables supervision of both masked and visible tokens during training, as well as self-correction during inference. We address the limitations of prior hollow transformers through two complementary advances: a new, lower-variance estimator of the MDM NELBO that supervises every position (Section~\ref{sec:method-objective}), and a redesigned merging mechanism that efficiently integrates left and right contexts at every layer (Sections~\ref{sec:method-architecture} and~\ref{sec:method-efficient}). Finally, we show how \method{}'s leave-one-out conditionals enable classical sampling techniques, including Gibbs-style and Metropolis--Hastings-style updates, allowing effective self-correction during generation (Section~\ref{sec:method-sampling}).

\subsection{Training Objective}
\label{sec:method-objective}

We exploit the leave-one-out property to derive an all-position estimator of the MDM NELBO. Unlike standard MDMs, which supervise only masked positions, \method{} can predict every token from its surrounding context. Formally, we require the following property.

\begin{definition}[Leave-one-out property]
\label{def:loo}
A denoiser $p_\theta$ satisfies the leave-one-out (LOO) property if its prediction at position $i$ depends on the input only through $\mathbf{x}_{-i}$. Equivalently, for any two inputs $\mathbf{x}$ and $\mathbf{x}'$ differing only at position $i$,
\[
    p_\theta(x_i = v \mid \mathbf{x})
    \;=\;
    p_\theta(x_i = v \mid \mathbf{x}')
    \quad \text{for all } v \in \mathcal{V}.
\]
\end{definition}

Crucially, the LOO property ensures that the model's prediction at position $i$ is the same whether the input at that position is masked or visible. For any mask set $M_t$ and position $i$, the following holds
\begin{equation}
\label{eq:loo-equality}
    p_\theta(x_i \mid \mathbf{x}[M_t])
    =
    p_\theta(x_i \mid \mathbf{x}[M_t \cup \{i\}]).
\end{equation}
This invariance allows us to analytically marginalize the masking probability at each position in the standard MDM NELBO, yielding the following objective:
\begin{equation}
\label{eq:loo-objective}
    \mathcal{L}_{\mathrm{LOOT}}(\theta)
    =
    \mathbb{E}_{t,\mathbf{x},M_t}
    \left[
        \alpha'_t
        \sum_{i=1}^{N}
        \log p_\theta\!\left(x_i \mid \mathbf{x}[M_t]\right)
    \right].
\end{equation}
The full derivation is provided in Appendix~\ref{app:loot-elbo}. This objective is a valid MDM likelihood bound whenever $p_\theta$ satisfies the LOO property. Every token contributes to the \method{} loss during training, yielding a lower-variance estimator of the MDM NELBO. Under the standard log-linear noise schedule $\alpha_t=1-t$ used by MDLM~\citep{sahoo2024mdlm} and LLaDA~\citep{nie2025largelanguagediffusionmodels}, the \method{} NELBO weights reduce to a constant $\alpha'_t=-1$ for all $t$. This eliminates the divergence of the standard MDLM NELBO weight $1/(1-\alpha_t)$, which diverges as $\alpha_t \to 1$.

\paragraph{Relation to dense MDM objectives.}
HollowDiff~\citep{zhao2024informed} and concurrently PGM~\citep{deschenaux2025partition} proposed the half-and-half loss, $\frac{1}{2}(\mathcal{L}_{M} + \mathcal{L}_{\bar{M}})$, which averages the MDM losses on masked and unmasked positions. While this provides supervision at every position, its NELBO weights diverge at both endpoints $\alpha_t \to 0$ and $\alpha_t \to 1$. Our objective is a Rao-Blackwellization of the half-and-half loss, analytically marginalizing the masking probability at each position and eliminating the weight divergence (Appendix~\ref{app:half-and-half}). In our ablations, we show that our proposed NELBO improves the validation perplexity on LM1B by $3.1$ for a \method{} model compared to the half-and-half loss (Table~\ref{tab:hollow_ablation}).

\subsection{The Leave-One-Out Transformer}
\label{sec:method-architecture}

Next, we propose a model architecture that efficiently evaluates $p_\theta(x_i \mid \mathbf{x}_{-i})$ at every position in a single forward pass, while preserving the LOO property~(Def. \ref{def:loo}) across layers.

\paragraph{Challenge.}
A single attention layer with mask $(\mathbf{1}\mathbf{1}^\top - I)$, ones everywhere except on the diagonal, satisfies Definition~\ref{def:loo}, but the property is lost under composition: a neighboring position $j$ attends to $x_i$ in one layer, and position $i$ attends to $j$'s updated representation in the next, routing $x_i$ back into its own prediction. Hollow transformers resolve this with two causal streams (Section~\ref{sec:bg-hollow}) which preserve the LOO property under composition.

\paragraph{Three Data Streams.} \method{} represents each sequence with three parallel streams of hidden states, illustrated in Figure~\ref{fig:loo_method}. (1) \textbf{Forward stream $\vec{\mathbf{f}}$:} encodes left context. At each layer, position $i$ attends to positions $\{1, \ldots, i\}$ via a standard left-to-right causal mask. (2) \textbf{Backward stream $\cev{\mathbf{b}}$:} encodes right context, mirrored: position $i$ attends to positions $\{i, \ldots, N\}$ via a right-to-left causal mask. These are the two causal transformers of \citet{sun2022score}. (3) \textbf{Output stream $\mathbf{o}$:} computes the leave-one-out conditional. At each layer, position $i$ attends to forward keys $\vec{\mathbf{f}}_{<i}$ and backward keys $\cev{\mathbf{b}}_{>i}$, merging bidirectional context that excludes $x_i$. This third stream follows HollowDiff~\citep{zhao2024informed}, with one important difference. \method{} merges at every layer rather than once or every few layers: bidirectional context accumulates with depth because each layer's contribution to $\mathbf{o}_i$ is added to the residual stream, and perplexity improves monotonically with merge frequency (Table~\ref{tab:hollow_ablation}).

\paragraph{Initialization.}
The forward and backward streams are initialized using the partially masked sequence $\mathbf{x}[M_t]$, while the output stream is initialized to the $\texttt{[MASK]}$ token embedding to ensure that predictions satisfy the LOO property. The output stream could alternatively be initialized using a separate learned embedding, which we did not explore in this work.

\begin{wrapfigure}{r}{0.42\textwidth}
    \vspace{-5em}
    \centering
    \includegraphics[width=\linewidth]{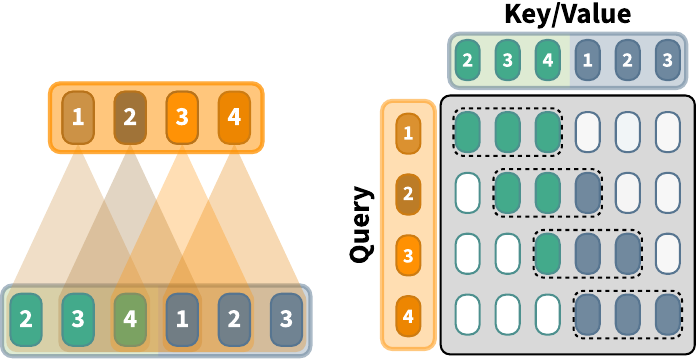}
    \caption{\textbf{Efficient Merger} using sliding-window attention.}
    \label{fig:sliding_window}
    \vspace{-1em}
\end{wrapfigure}

\subsection{An Efficient Implementation of \method{}}
\label{sec:method-efficient}

\paragraph{Parameterization.}
Aside from the attention masks, \method{} follows the standard DiT-style MDM architecture \citep{peebles2023scalablediffusionmodelstransformers, sahoo2024mdlm}, with adaptive layer norms, MLPs, and residual connections (full details in Appendix~\ref{app:training-architectures}). The three streams share all attention projection matrices ($W_Q$, $W_K$, $W_V$, $W_O$) at each layer, so the parameter count of \method{} matches that of a standard DiT of the same depth and width. The forward and backward streams additionally share layer norms; the output stream has its own. This minor parameter overhead gives the output stream, which solves a structurally different task than the directional streams, additional expressivity at low cost.

\paragraph{Efficient Merging with FlashAttention.}

The merge block's attention pattern admits a clean reformulation as sliding-window attention. By concatenating the backward and forward streams along the sequence dimension to form the key-value tensor $\mathbf{bf} = [\cev{\mathbf{b}}_{2:N}, \vec{\mathbf{f}}_{1:N-1}]$, each output query $\mathbf{o}_i$ attends to a contiguous length-$(N-1)$ window $[i, i + N - 2]$ of $\mathbf{bf}$ (Figure~\ref{fig:sliding_window}). This reduces the merge to a standard sliding-window attention pattern, executable in a single FlashAttention call~\citep{dao2022flashattention}. The forward and backward streams have identical mask structure (causal in their respective directions) and are batched into a single FlashAttention call, so each \method{} layer requires only two FlashAttention calls in total, with no custom kernels.

\paragraph{Computational Overhead.} Counting matmul and attention FLOPs at $N = 1024$, a \method{} forward pass costs $2.24\times$ that of a standard MDM of the same size (Appendix~\ref{app:computational-cost}). This is well below the naive $3\times$ of three streams: \method{} doubles rather than triples the attention work, its final merge block is cheaper than a full layer, and the vocabulary head is shared. With \texttt{torch.compile}, the forward+backward cost ratio ranges from $1.25\times$ at small batch sizes, where both models are kernel-launch-bound, to $2.22\times$ at large batch sizes and $N=1024$, where both are compute-bound. This per-pass overhead is repaid at inference, where a single \method{} pass returns all $N$ leave-one-out conditionals (Section~\ref{sec:method-sampling}).

\subsection{Self-correction and sampling}
\label{sec:method-sampling}

\begin{wrapfigure}{r}{0.48\textwidth}
    \vspace{-2em}
    \centering
    \includegraphics[width=0.48\textwidth]{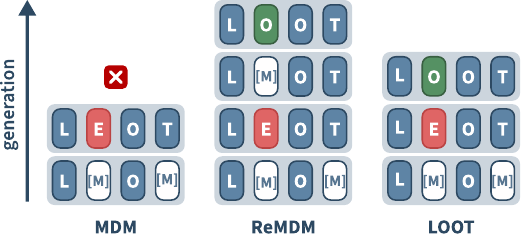}
\caption{\method{} enables efficient self-correction.}
\vspace{-2em}
\label{fig:repair_illustration}
\end{wrapfigure}

A trained \method{} model predicts the full set of leave-one-out conditionals $\{p_\theta(x_i \mid \mathbf{x}_{-i})\}_{i=1}^N$ in a single forward pass. These conditionals provide a distribution over replacements at each token position, enabling the use of classical sampling algorithms, such as Gibbs- and Metropolis-style methods. We outline some of the self-correction procedures, which enable efficient and effective self-correction during inference.

\paragraph{Two-stage sampling.}
We generate a sample in two stages. First, we produce an initial sequence $\mathbf{x}^{(0)}$ using $S_{\text{init}}$ standard unmasking steps. Second, we run $R$ self-correction steps: at each step, one \method{} forward pass returns $p_\theta(x_i \mid \mathbf{x}^{(r)}_{-i})$ for every position $i$, and a correction policy selects positions to edit and resamples them. The total number of forward passes is $\text{NFE} = S_{\text{init}} + R$.

\paragraph{Correction policies.}
\method{}'s leave-one-out conditionals admit several principled correction policies; we summarize the main ones here and defer full details to Appendix~\ref{app:self-correction-policies}.

The simplest is \emph{blocked Gibbs sampling}: partition positions into $G$ groups, and at each correction step, resample one entire group in parallel from $\{p_\theta(x_i \mid \mathbf{x}_{-i})\}_{i \in G_r}$.

A second policy is \emph{Metropolis-Hastings}, which proposes a replacement $z_i \sim p_\theta(\cdot \mid \mathbf{x}_{-i})$ and accepts it with probability $\alpha_i = \min(1,\, p_\theta(z_i \mid \mathbf{x}_{-i}) / p_\theta(x_i \mid \mathbf{x}_{-i}))$, comparing the model's preference for the proposed token against its preference for the current one. If the proposal is more compatible with the surrounding context than the current token, it is accepted with probability one; if less compatible, the edit is accepted only stochastically, preventing the sampler from overwriting good tokens just because a sample happened to land on something different.

Finally, \emph{improvement-based} policies score each position by the log-likelihood gap $\Delta_i = \max_{v \neq x_i} \log p_\theta(v \mid \mathbf{x}_{-i}) - \log p_\theta(x_i \mid \mathbf{x}_{-i})$ between the best alternative replacement and the current token. Positions with $\Delta_i > 0$ are \emph{locally improvable}: under the model, some other token fits the surrounding context better than the current one. Greedy variants edit the position with the largest $\Delta_i$; stochastic variants sample positions with probability proportional to $\Delta_i$.

\paragraph{Comparison to remasking.}
Remasking-based methods such as ReMDM~\citep{wang2025remasking} require multiple forward passes per correction (Figure~\ref{fig:repair_illustration}): one to remask a token, and another to resample. Worse, choosing \emph{which} token to remask requires the up-to-date conditionals $p_\theta(x_i \mid \mathbf{x}_{-i})$, which a standard MDM cannot evaluate efficiently: recovering all $N$ would take $N - |M_t|$ forward passes, masking each unmasked position individually. ReMDM falls back to heuristics like confidence at the time of unmasking, which grow stale as context fills in. \method{} returns all $N$ conditionals in a single forward pass, so each correction step is one model evaluation regardless of how many tokens are revised.

\section{Experiments}

\begin{figure*}[t]
    \centering
    \begin{subfigure}[t]{0.56\textwidth}
        \centering
        \includegraphics[width=\linewidth]{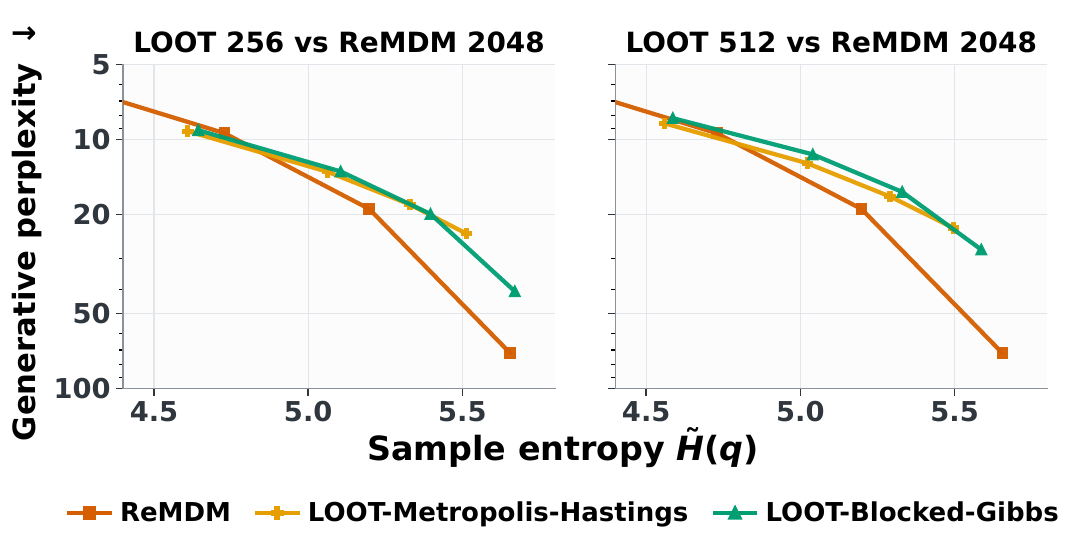}
        \caption{Unequal-compute frontier.}
        \label{fig:sampling-compute-efficiency}
    \end{subfigure}
    \hfill
    \begin{subfigure}[t]{0.4\textwidth}
        \centering
        \includegraphics[width=\linewidth]{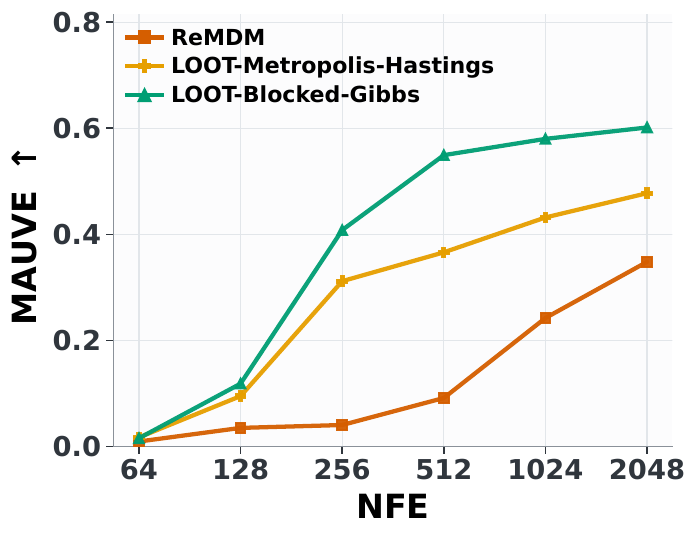}
        \caption{MAUVE across NFE budgets.}
        \label{fig:sampling-mauve-vs-nfe}
    \end{subfigure}
    \caption{\textbf{\method{} self-correction outperforms ReMDM with $8\times$ fewer sampling steps.}
\textbf{(a)} Quality--diversity frontiers on OpenWebText across sampling temperatures. At 256 and 512 NFEs, \method{} matches or surpasses ReMDM at 2048 NFEs across the frontier, while using $3.5\times$ and $1.7\times$ fewer inference FLOPs, respectively. \textbf{(b)} MAUVE ($\uparrow$) across NFE budgets at matched entropy. \method{}-Blocked-Gibbs at 256 NFEs surpasses ReMDM at 2048 NFEs.
    }
    \vspace{-1.25em}
    \label{fig:sampling-compute-summary}
\end{figure*}

\subsection{Language Modeling}
\label{sec:val-ppl-results}

\paragraph{Setup.}
We evaluate \method{} against discrete diffusion baselines (Table~\ref{tab:main_ppl}) on two standard language modeling datasets: LM1B~\citep{lm1b2014}, with sequence length $N=128$ and 1M training steps, and OpenWebText~\citep{Gokaslan2019OpenWeb} (OWT), with $N=1024$ and 250K training steps. We follow the training setup of \citet{sahoo2026esolm}, matching model parameter counts across diffusion baselines within each benchmark. Following standard practice~\citep{sahoo2024mdlm}, we report diffusion perplexities using the NELBO upper bound on negative log-likelihood. Full training details are provided in Appendix~\ref{app:training-details}.

\begin{table}[b!]
\vspace{-1.25em}
\centering
\caption{Validation perplexity (PPL) on LM1B ($N=128$, 1M steps) and OWT ($N=1024$, 250K steps). MDLM, PGM, and \method{} report mean $\pm$ std over 5 evaluation seeds.
$^{\diamond}$Reported in Sahoo et al.~\citep{sahoo2026esolm}.
$^{\ddagger}$Reproduced from checkpoints provided by Sahoo et al.~\citep{sahoo2024mdlm,sahoo2025duo} and Schiff et al.~\citep{schiff2025udlm} in~\citep{sahoo2026esolm}.
$^{\dagger}$Trained for 1M OWT steps (not 250K).
$^{\P}$No sentence packing.
\label{tab:main_ppl}}
\newlength{\stdwidth}
\newlength{\numwidth}
\newlength{\pplpad}
\newlength{\stdover}
\newcommand{\stdfmt}[1]{${}_{\scriptscriptstyle\pm#1}$}
\newcommand{\std}[1]{\rlap{\textcolor{black!50}{\stdfmt{#1}}}}
\newcolumntype{P}{>{\makebox[\pplpad]{}}r<{\makebox[\stdwidth]{}}}
\newcommand{\pplhead}{\multicolumn{1}{P}{\makebox[\numwidth]{\textbf{PPL ($\downarrow$)}}}}

\definecolor{lightpurple}{RGB}{230, 220, 245}
\definecolor{lightblue}{RGB}{220, 235, 250}

{\setlength{\tabcolsep}{4pt}
\settowidth{\stdwidth}{\stdfmt{0.00}}
\settowidth{\numwidth}{00.00}
\settowidth{\pplpad}{\textbf{PPL ($\downarrow$)}}
\setlength{\pplpad}{\dimexpr(\pplpad-\numwidth)/2\relax}
\setlength{\stdover}{\dimexpr\stdwidth-\pplpad\relax}
\begin{tabular}{llcPcP}
\toprule
\textbf{Category} & \textbf{Method}
& \multicolumn{2}{c}{\textbf{LM1B} ($N=128$)\makebox[\stdover]{}}
& \multicolumn{2}{c}{\textbf{OWT} ($N=1024$)\makebox[\stdover]{}} \\
\cmidrule(lr{\dimexpr\stdover+\cmidrulekern\relax}){3-4}
\cmidrule(lr{\dimexpr\stdover+\cmidrulekern\relax}){5-6}
& & \textbf{Params} & \pplhead
& \textbf{Params} & \pplhead \\
\midrule

Autoregressive
& Transformer + AdaLN$^{\diamond}$~\citep{sahoo2026esolm}
& 132M & 21.86
& 162M & 17.78\rlap{$^{\dagger}$} \\
\midrule

\multirow{5}{*}{Diffusion}
& SEDD Absorb$^{\ddagger}$~\citep{lou2024sedd}
& 139M & 32.71\rlap{$^{\P}$}
& 170M & 26.81 \\

& MDLM$^{\ddagger}$~\citep{sahoo2024mdlm}
& 139M & 31.80\std{0.08}
& 170M & 25.20\std{0.02} \\

& UDLM$^{\ddagger}$~\citep{schiff2025udlm}
& 139M & 36.71
& 170M & 30.52 \\

& Duo$^{\ddagger}$~\citep{sahoo2025duo}
& 139M & 33.68
& 170M & 27.14 \\

& PGM~\citep{deschenaux2025partition}
& 142M & 31.04\std{0.09}
& 171M & 26.57\std{0.02} \\

\midrule
\rowcolor{blue!10}
Ours & \method{}
& 139M & 31.02\std{0.04}
& 170M & 26.83\std{0.01} \\
\bottomrule
\end{tabular}}

\end{table}

\paragraph{Likelihood evaluation.}
On LM1B, \method{} improves over MDLM by $0.78$ perplexity and matches PGM~\citep{deschenaux2025partition}, the strongest prior dense-supervision MDM baseline, with differences between the two within evaluation-seed variation. On OWT at $N=1024$, \method{} achieves a perplexity of $26.83$, compared with $25.20$ for MDLM and $26.57$ for PGM. Thus, while \method{} is competitive with standard MDMs, a modest performance gap remains at longer sequence lengths.

We conduct two ablations to investigate the likelihood gap at longer sequence lengths and evaluate our improvements over prior hollow transformers (Appendix~\ref{app:ablations}).
\vspace{-0.5em}
\paragraph{Sequence length ablations.}
We train three models on OWT at sequence lengths $N=128$ and $N=1024$, holding all other settings fixed (Table~\ref{tab:length_ablation}): standard MDLM, the \method{} architecture trained with the MDLM loss, and \method{} trained with our all-position objective. Fixing the \method{} architecture, our objective improves perplexity by $2.9$ and $2.8$ at $N=128$ and $N=1024$, respectively. This consistent improvement allows \method{} to outperform MDLM at $N=128$ on both OWT and LM1B. However, holding the MDLM loss fixed reveals a gap: the leave-one-out architecture incurs a perplexity penalty of $2.2$ at $N=128$, increasing to $4.4$ at $N=1024$. Thus, while our objective more than compensates for this penalty at shorter sequences, it cannot fully close the gap at longer sequences. These findings suggest a length-dependent modeling capacity gap between hollow and standard bidirectional transformers. Nevertheless, \method{} achieves better perplexity than Duo~\citep{sahoo2025duo} and UDLM~\citep{schiff2025udlm}, two diffusion language models that also support native self-correction.
\vspace{-0.5em}
\paragraph{Hollow transformer ablations.}
Table~\ref{tab:hollow_ablation} evaluates our architectural and training improvements using shorter LM1B runs. First, more frequent merging of left and right contexts consistently improves perplexity, with per-layer merging achieving the best performance. Second, our \method{} NELBO provides substantial gains independently of merging frequency: it improves perplexity over the half-and-half loss used by prior works~\citep{zhao2024informed, deschenaux2025partition}. Notably, even the merge-once architecture of \citet{sun2022score} trained with our objective approaches the bidirectional DiT baseline ($41.07$ vs.\ $39.80$), while per-layer merging surpasses it ($36.23$ vs.\ $39.80$). These results demonstrate that both our objective and architecture contribute to making hollow transformers competitive language models.

\subsection{Sampling and Self-correction}
\label{sec:sampling-finetuning}

\paragraph{Setup.} We evaluate unconditional generation on OpenWebText (OWT) at sequence length $N=1024$. We compare \method{} and MDLM trained from scratch for 250K steps on the same data, with global batch size 512 and the same log-linear noise schedule (Appendix~\ref{app:training-details}). We evaluate MDLM~\citep{sahoo2024mdlm}, ReMDM-loop~\citep{wang2025remasking}, and \method{} with and without self-correction. Compute is measured in number of forward evaluations (NFE): for MDLM and ReMDM-loop, the number of reverse denoising steps; for \method{}, $S_{\text{init}} + R$, the sum of initial denoising steps and refinement steps.

\paragraph{Evaluation protocol.}
We follow \citet{pynadath2026generativefrontiersevaluationmatters} and report the trade-off between generative perplexity (lower is better, a quality measure) and sample entropy $\tilde{H}(q)$ (higher is better, a diversity measure). We use GPT-2 Large~\citep{radford2019language} to evaluate generative perplexity and generate 512 samples of length $N=1024$ per configuration for the frontiers. We additionally report MAUVE~\citep{pillutla2021mauve}, which captures both generation quality and cross-sample diversity, using 5,000 samples at temperatures matched to the entropy of OWT validation text. Following ReMDM~\citep{wang2025remasking}, we use nucleus sampling with $p=0.9$, but include quality--diversity frontiers for $p=1.0$ in Appendix~\ref{app:topp1}. The full sampling evaluation details are provided in Appendix~\ref{app:sampling-eval-details}.

\paragraph{Compute efficiency.}

\begin{table*}[h!]
    \centering
    \footnotesize
    \caption{\textbf{Unconditional generation quality on OpenWebText.}
    We compare \method{} sampling methods against MDLM and ReMDM
    across different numbers of function evaluations (NFEs). \method{}-Blocked-Gibbs achieves better generation quality than ReMDM with $8\times$ fewer NFEs (256 vs.\ 2048) at matched entropy.}
    \label{tab:owt-generation}

    \setlength{\tabcolsep}{4pt}
    \begin{tabular}{lrcccc|cccc|cccc}
    \toprule
    & & \multicolumn{4}{c|}{MAUVE ($\uparrow$)}
        & \multicolumn{4}{c|}{Gen. PPL ($\downarrow$)}
        & \multicolumn{4}{c}{Entropy} \\
    \midrule
    \rowcolor{lightgray!25}
    \multicolumn{2}{r}{$\mathrm{NFE} =$}
    & 256 & 512 & 1024 & 2048
    & 256 & 512 & 1024 & 2048
    & 256 & 512 & 1024 & 2048 \\

    \multicolumn{2}{l}{MDLM}
    & 0.014 & 0.015 & 0.017 & 0.018
    & 63.1 & 60.8 & 59.4 & 58.5
    & 5.53 & 5.52 & 5.51 & 5.50 \\

    \multicolumn{2}{l}{ReMDM}
    & 0.040 & 0.091 & 0.242 & 0.348
    & 53.9 & 44.9 & 37.6 & 31.2
    & 5.54 & 5.50 & 5.45 & 5.40 \\

    \midrule
    \multicolumn{2}{l}{\method{}-Baseline}
    & 0.010 & 0.010 & 0.012 & 0.012
    & 55.4 & 50.8 & 48.8 & 47.6
    & 5.47 & 5.45 & 5.43 & 5.42 \\

    \rowcolor{blue!10}
    \multicolumn{2}{l}{\method{}-Blocked-Gibbs}
    & \textbf{0.408} & \textbf{0.549}
    & \textbf{0.580} & \textbf{0.602}
    & 20.1 & 20.0 & 21.1 & 17.5
    & 5.40 & 5.45 & 5.51 & 5.46 \\

    \multicolumn{2}{l}{\method{}-Best-Improvement}
    & 0.035 & 0.072 & 0.201 & 0.401
    & 36.3 & 32.3 & 29.2 & 24.3
    & 5.44 & 5.48 & 5.52 & 5.48 \\

    \multicolumn{2}{l}{\method{}-Metropolis-Hastings}
    & 0.312 & 0.366 & 0.432 & 0.478
    & 21.2 & 22.8 & 22.1 & 21.8
    & 5.43 & 5.49 & 5.47 & 5.47 \\

    \bottomrule
    \end{tabular}
\end{table*}

Figure~\ref{fig:sampling-compute-summary} highlights our main result: \method{} achieves favorable generation quality with substantially less inference compute. Figure~\ref{fig:sampling-compute-summary}(a) shows that \method{} at 256 and 512 NFEs matches or surpasses ReMDM at 2048 NFEs across the quality--diversity frontier. More strikingly, Figure~\ref{fig:sampling-compute-summary}(b) shows that \method{}-Blocked-Gibbs at just 256 NFEs achieves a MAUVE score of $0.408$, exceeding ReMDM's $0.348$ at 2048 NFEs, with matched sample entropy. These results correspond to $8\times$ fewer forward evaluations and $3.5\times$ fewer inference FLOPs, accounting for the $2.2\times$ cost of a \method{} forward pass. At 512 NFEs, \method{} further improves both metrics while requiring $1.7\times$ fewer inference FLOPs. In wall-clock time, \method{} at 256 NFEs is approximately $8\times$ faster than ReMDM at 2048 NFEs on the same hardware (Table~\ref{tab:inference-cost}; Appendix~\ref{app:computational-cost}).

\paragraph{Quality--diversity frontier.}

\begin{figure*}[t]
    \centering
    \includegraphics[width=\textwidth]{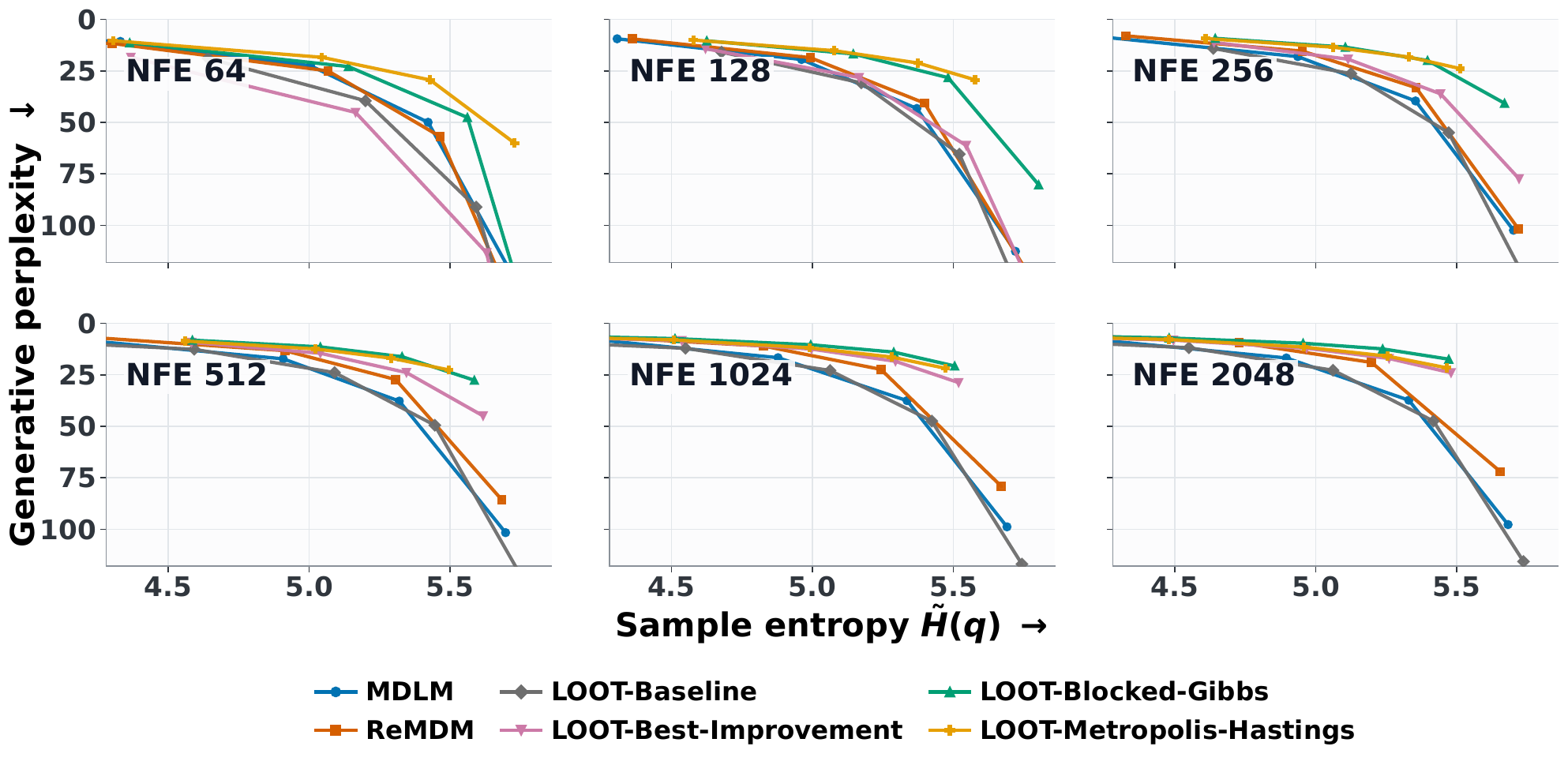}
    \caption{\textbf{\method{} Self-Correction Improves the Sampling Frontier.}
    Generative perplexity versus sample entropy on OpenWebText across NFE budgets.
    \method{} self-correction improves the frontier relative to MDLM, ReMDM-loop,
    and the uncorrected \method{} baseline.}
    \label{fig:sampling-frontier-all-nfe}
\end{figure*}

Figure~\ref{fig:sampling-frontier-all-nfe} shows that \method{} with self-correction consistently outperforms ReMDM across the full quality--diversity frontier at every evaluated NFE budget. This demonstrates that \method{} achieves a better tradeoff between generation quality and diversity.

\paragraph{MAUVE.}
MAUVE confirms these improvements at the distribution level (Table~\ref{tab:owt-generation}). At just 256 NFEs, \method{}-Blocked-Gibbs achieves MAUVE $0.408$, surpassing ReMDM's $0.348$ at 2048 NFEs, with matched entropy. Crucially, these gains come from self-correction: \method{}-Baseline, which uses standard MDM ancestral sampling without self-correction, achieves only $0.010$ MAUVE at 256 NFEs, even below MDLM ($0.014$). Applying Blocked Gibbs or Metropolis--Hastings correction to the same model raises MAUVE to $0.408$ and $0.312$, respectively. However, not all correction strategies perform equally well: Best Improvement reaches only $0.035$. These results highlight the importance of the sampling algorithm and suggest substantial room for improvement beyond the three classical correction methods explored here. Full MAUVE results are provided in Appendix~\ref{app:mauve-full}.

\section{Discussion}

\paragraph{Disentangling architecture, training, and self-correction.}
Our ablations show that the \method{} architecture improves likelihood over prior hollow transformers (Table~\ref{tab:hollow_ablation}), while the \method{} loss further improves validation perplexity over both the standard MDM loss (Table~\ref{tab:length_ablation}) and the half-and-half loss (Table~\ref{tab:hollow_ablation}). We further observe that the \method{} architecture enables effective self-correction independently of the training loss (Appendix~\ref{app:hybrid}). Even when trained with the standard masked-only MDM loss, \method{} substantially improves generation quality through self-correction, matching or surpassing ReMDM despite performing considerably worse without correction (Table~\ref{tab:hybrid}). Together, these results establish the distinct contributions of our architecture and loss: the architectural redesign improves hollow language modeling and enables effective self-correction, while the \method{} loss further improves language modeling performance and training efficiency.

\paragraph{Compatibility with pretrained checkpoints.}
An important practical question is whether \method{} can leverage existing pretrained models. Starting from an MDM checkpoint pretrained for 200K steps, we compare continuing standard MDM training for 50K steps against converting the checkpoint to \method{} and finetuning with our objective for the same duration (Appendix~\ref{app:finetune}). Both models are trained on the same number of tokens, but the \method{} finetune incurs approximately $2.2\times$ the per-step FLOPs, resulting in $1.25\times$ the total training FLOPs. We find that finetuned \method{} retains comparable generation quality without correction, while self-correction yields substantial improvements over continued MDM training. This demonstrates that existing MDM checkpoints can be adapted for self-correcting generation without retraining from scratch.

\paragraph{Hybrid training and inference.}
Full parameter sharing allows \method{} to recover standard bidirectional attention (Appendix~\ref{app:compatibility}). This opens the possibility of hybrid training and inference, using the cheaper MDM mode for highly masked sequences and switching to leave-one-out prediction for dense supervision and self-correction as more tokens become visible. The causal streams also suggest potential initialization from pretrained autoregressive models, particularly for block diffusion~\citep{arriola2025blockdiffusioninterpolatingautoregressive}.

\paragraph{Limitations and future directions.}
Despite these improvements, \method{} retains a likelihood gap relative to standard bidirectional MDMs at longer sequence lengths and incurs approximately $2.2\times$ the FLOPs per forward pass. Moreover, the substantial differences between our three self-correction strategies demonstrate that generation quality depends strongly on the sampling algorithm. Understanding the source of the length-dependent likelihood gap, developing more effective self-correction methods, and exploring hybrid training and inference remain promising directions for future work.

\section{Related Work}
\label{sec:related}

\paragraph{Diffusion Language Models.}
Masked diffusion language models were introduced by D3PM~\citep{austin2023structureddenoisingdiffusionmodels} and refined by SEDD~\citep{lou2024sedd}, MDLM~\citep{sahoo2024mdlm}, and MD4~\citep{shi2025simplifiedgeneralizedmaskeddiffusion}. Subsequent work explored hybrid AR--diffusion factorizations~\citep{arriola2025blockdiffusioninterpolatingautoregressive, lovelace2026stop, sahoo2026esolm, arriola2026set} and large-scale pretraining~\citep{nie2025largelanguagediffusionmodels, bie2025llada20scalingdiffusionlanguage, ye2025dream7bdiffusionlarge, fu2026nemotron}. A parallel line of work revisits noising processes that permit token revision during denoising~\citep{schiff2025udlm, von2025generalized, sahoo2025duo}. While these objectives expose self-correction directly through the noise process, they remain empirically harder to optimize in parameter-matched settings~\citep{sahoo2026scaling}. \method{} instead retains the standard absorbing-state objective and introduces self-correction through the architecture.

\paragraph{Hollow transformers and leave-one-out denoisers.}
Hollow transformers were introduced by \citet{sun2022score} to compute the singleton conditionals required by ratio matching in a single forward pass. HollowDiff~\citep{zhao2024informed} extended this architecture for diffusion-based correctors, while related constructions have been explored in discrete neural samplers~\citep{ou2026discrete}. We discuss these architectures and their limitations in Section~\ref{sec:bg-hollow}. In concurrent work, \citet{gourevitch2026uniform} show that leave-one-out prediction arises naturally in uniform diffusion and can be obtained through a parameterization without architectural changes. In contrast, under masked noising, standard bidirectional transformers receive no supervision for leave-one-out predictions at visible positions. \method{} therefore enforces the property architecturally, allowing every position to be predicted from its surrounding context.

\paragraph{Inference-time self-correction.}
Several methods seek to address the inability of standard MDMs to revise previously generated tokens. ReMDM~\citep{wang2025remasking} enables self-correction through remasking, but relies on heuristics to determine which tokens to remask. Other approaches introduce auxiliary correction objectives that explicitly train models to recognize and correct errors~\citep{kim2025fine, schiff2026learnmistakesselfcorrectingmasked}, departing from the standard variational likelihood objective. In contrast, \method{} directly models the leave-one-out conditionals $p_\theta(x_i \mid \mathbf{x}_{-i})$ at every position in a single forward pass, enabling principled Gibbs-style and Metropolis-style self-correction without heuristic remasking or auxiliary correction losses, while retaining training under the MDM NELBO.

\paragraph{Dense supervision of MDMs.}
PGM~\citep{deschenaux2025partition} improves training and sampling efficiency by partitioning tokens into masked and unmasked subsets, using restricted attention to prevent information leakage between them. However, PGM, like standard MDMs, cannot efficiently revise previously generated tokens. In contrast, \method{} models the leave-one-out conditionals $p_\theta(x_i \mid \mathbf{x}_{-i})$ at every position, enabling both dense supervision during training and self-correction during generation. Our Rao--Blackwellized NELBO estimator further reduces training variance while preserving the standard absorbing-state diffusion objective. We compare directly against PGM in our experiments, and provide a more detailed discussion of the loss in Appendix~\ref{app:half-and-half}.

\section{Conclusion}

We introduce the Leave-One-Out Transformer (\method{}), a more expressive and efficient hollow architecture. By predicting every token from its surrounding context, \method{} admits a lower-variance training objective that supervises all positions, as well as Gibbs-style sampling methods for self-correction. A single \method{} forward pass computes all leave-one-out conditionals, whereas a standard MDM would require $N-|M_t|$ separate forward passes to obtain these predictions for visible tokens. Although each \method{} forward pass requires $2.2\times$ the FLOPs of a standard MDM, it enables high-quality, few-step generation. On OpenWebText, \method{} achieves better generation quality than remasking-based methods with $8\times$ fewer forward evaluations and $3.5\times$ fewer inference FLOPs. Future work may explore more effective self-correction algorithms and hybrid training and inference schemes that combine the efficiency of standard MDMs with the refinement capabilities of \method{}.

\section{Acknowledgments}
We thank Eric Enouen and Linxi Zhao for insightful discussions, help with the figures, and valuable feedback on the manuscript. This work was funded by NSF CAREER Award No. 2046760 and NSF grant IIS-2530143. This material is based on work supported by the AI Research Institutes program supported by the NSF and Intel Corporation under NSF award DMR-2433348.

\bibliography{refs}

\clearpage
\allowdisplaybreaks
\appendix
\setcounter{tocdepth}{2}
\tableofcontents
\newpage
\section{Derivation of the LOOT Training Objective}
\label{app:loot-elbo}

This appendix derives the LOOT training objective from the masked diffusion NELBO under the leave-one-out (LOO) property of the denoiser (Definition~\ref{def:loo}). The argument is direct: under the LOO property, the per-position expectation in the MDM NELBO admits a Rao-Blackwellization that absorbs the $1/(1 - \alpha_t)$ weighting and yields a sum over all positions rather than a sum over the random subset $M_t$. We then discuss the connection to the half-and-half objective used by HollowDiff~\citep{zhao2024informed} and PGM~\citep{deschenaux2025partition}, which arises as an intermediate form and exhibits known numerical instability.

\paragraph{Setup.} Let $\mathbf{x} \sim p_{\text{data}}$ be a sequence of length $N$ over a vocabulary $\mathcal{V}$ that includes a dedicated $\texttt{[MASK]}$ token. The continuous-time forward process samples a mask $M_t \subseteq \{1, \ldots, N\}$ at time $t \in [0, 1]$, where each position is masked independently with probability $1 - \alpha_t$. We write $\mathbf{x}[M_t]$ for the sequence with positions in $M_t$ replaced by $\texttt{[MASK]}$, and assume a time-independent denoiser $p_\theta(x_i \mid \mathbf{x}[M_t], t) = p_\theta(x_i \mid \mathbf{x}[M_t])$ throughout, following standard practice in MDM~\citep{sahoo2024mdlm, shi2025simplifiedgeneralizedmaskeddiffusion}. The schedule satisfies $\alpha_0 = 1$ and $\alpha_1 = 0$, so $\alpha_t' < 0$.

The standard MDM continuous-time NELBO is
\begin{equation}
\label{eq:mdm-nelbo-app}
\mathcal{L}_{\text{MDM}}(\theta) = \mathbb{E}_{\mathbf{x}} \!\left[ \int_0^1 \frac{\alpha_t'}{1 - \alpha_t} \, \mathbb{E}_{M_t}\!\left[ \sum_{i \in M_t} \log p_\theta(x_i \mid \mathbf{x}[M_t]) \right] dt \right].
\end{equation}
The loss is supported only on positions $i \in M_t$, and the $\alpha_t' < 0$ factor combined with $\log p_\theta < 0$ yields a positive loss.

\subsection{The LOOT NELBO}
\label{app:loot-nelbo-math}

Under the LOO property (Definition~\ref{def:loo}), masking position $i$ does not change the prediction at that position:
\begin{equation}
\label{eq:loo-equiv-app}
p_\theta(x_i \mid \mathbf{x}[M_t])
=
p_\theta(x_i \mid \mathbf{x}[M_t \cup \{i\}])
\qquad \text{for all } i, M_t.
\end{equation}
We exploit this invariance to rewrite the MDM NELBO as a sum over all positions and marginalize each position's own masking event.

\paragraph{Rewriting as a sum over all positions.}
The masked-position sum in $\mathcal{L}_{\text{MDM}}$ can be expressed using indicator variables:
\begin{equation}
\label{eq:mdm-as-indicator}
\sum_{i \in M_t}
\log p_\theta(x_i \mid \mathbf{x}[M_t])
=
\sum_{i=1}^{N}
\mathbf{1}\{i \in M_t\}
\log p_\theta(x_i \mid \mathbf{x}[M_t]).
\end{equation}
Substituting into \eqref{eq:mdm-nelbo-app} and exchanging summation and expectation gives
\begin{equation}
\label{eq:mdm-rewritten}
\mathcal{L}_{\text{MDM}}(\theta)
=
\mathbb{E}_{\mathbf{x}}
\left[
\int_0^1
\frac{\alpha_t'}{1-\alpha_t}
\sum_{i=1}^{N}
\mathbb{E}_{M_t}
\left[
\mathbf{1}\{i \in M_t\}
\log p_\theta(x_i \mid \mathbf{x}[M_t])
\right] dt
\right].
\end{equation}

\paragraph{Rao--Blackwellization.}
Fix a position $i$ and let
$M_t^{(-i)} := M_t \setminus \{i\}$.
We Rao--Blackwellize the $i$-th summand by conditioning on $M_t^{(-i)}$:
\begin{align}
&\mathbb{E}_{M_t}
\left[
\mathbf{1}\{i \in M_t\}
\log p_\theta(x_i \mid \mathbf{x}[M_t])
\right]
\nonumber\\
&\quad=
\mathbb{E}_{M_t^{(-i)}}
\left[
\mathbb{E}
\left[
\mathbf{1}\{i \in M_t\}
\log p_\theta(x_i \mid \mathbf{x}[M_t])
\;\middle|\;
M_t^{(-i)}
\right]
\right].
\end{align}
By the LOO property, the prediction at position $i$ is unchanged whether that position is masked or visible. Consequently,
$\log p_\theta(x_i \mid \mathbf{x}[M_t])$
depends only on $M_t^{(-i)}$ and can be pulled out of the inner expectation:
\begin{align}
&\mathbb{E}
\left[
\mathbf{1}\{i \in M_t\}
\log p_\theta(x_i \mid \mathbf{x}[M_t])
\;\middle|\;
M_t^{(-i)}
\right]
\nonumber\\
&\quad=
\log p_\theta
\left(x_i \mid
\mathbf{x}[M_t^{(-i)} \cup \{i\}]
\right)
\Pr(i \in M_t \mid M_t^{(-i)})
\nonumber\\
&\quad=
(1-\alpha_t)
\log p_\theta
\left(x_i \mid
\mathbf{x}[M_t^{(-i)} \cup \{i\}]
\right),
\end{align}
where the final equality follows from the independence of per-position masking.

\paragraph{Cancellation and final form.}
Substituting into \eqref{eq:mdm-rewritten}, the factor $1-\alpha_t$ obtained by marginalizing the mask indicator cancels the inverse masking-probability weight:
\begin{align}
\mathcal{L}_{\text{MDM}}(\theta)
&=
\mathbb{E}_{\mathbf{x}}
\left[
\int_0^1
\frac{\alpha_t'}{1-\alpha_t}
\sum_{i=1}^{N}
\mathbb{E}_{M_t^{(-i)}}
\left[
(1-\alpha_t)
\log p_\theta
\left(x_i \mid
\mathbf{x}[M_t^{(-i)} \cup \{i\}]
\right)
\right]dt
\right]
\nonumber\\
&=
\mathbb{E}_{\mathbf{x}}
\left[
\int_0^1
\alpha_t'
\sum_{i=1}^{N}
\mathbb{E}_{M_t^{(-i)}}
\left[
\log p_\theta
\left(x_i \mid
\mathbf{x}[M_t^{(-i)} \cup \{i\}]
\right)
\right]dt
\right].
\end{align}
Applying the LOO property once more, the log-probability is unchanged whether position $i$ is masked or visible. We can therefore replace the expectation over $M_t^{(-i)}$ with an expectation over the full mask set $M_t$, yielding
\begin{equation}
\label{eq:loo-objective-app}
\boxed{
\mathcal{L}_{\text{LOOT}}(\theta)
=
\mathbb{E}_{t,\mathbf{x},M_t}
\left[
\alpha_t'
\sum_{i=1}^{N}
\log p_\theta(x_i \mid \mathbf{x}[M_t])
\right].
}
\end{equation}
This is the objective used in the main text (Equation~\ref{eq:loo-objective}). Unlike the standard MDM loss, the LOOT loss supervises every position while preserving the same expected NELBO.

\paragraph{Properties.}
The LOOT NELBO has three useful properties:
\begin{enumerate}[leftmargin=*, itemsep=2pt]
    \item \textbf{Equality with the MDM NELBO.}
    The Rao-Blackwellization preserves the expected value, so
    $\mathcal{L}_{\text{LOOT}}(\theta)
    = \mathcal{L}_{\text{MDM}}(\theta)$
    for any denoiser satisfying Definition~\ref{def:loo}.
    Therefore, the \method{} objective is a valid likelihood bound.

        \item \textbf{Lower variance than $\mathcal{L}_{\text{MDM}}$.}
    Fix $(\mathbf{x},t)$ and write
    $\ell_i := \log p_\theta(x_i \mid \mathbf{x}[M_t])$
    and $b_i := \mathbf{1}\{i \in M_t\}$.
    By the LOO property, $\ell_i$ is independent of $b_i$, so every term of the MDM estimator has larger variance than the corresponding LOOT term, at every noise level:
    \begin{equation}
    \label{eq:per-position-variance}
    \mathrm{Var}\left(
    \frac{b_i\ell_i}{1-\alpha_t}
    \right)
    =
    \mathrm{Var}(\ell_i)
    +
    \frac{\alpha_t}{1-\alpha_t}\,
    \mathbb{E}[\ell_i^2].
    \end{equation}
    The excess diverges as $\alpha_t \to 1$. Consequently, under the log-linear schedule and with bounded log-probabilities, the variance of the single-sample MDM estimator grows logarithmically as the time-clipping threshold approaches zero whenever the loss does not vanish at low noise, whereas the variance of the LOOT estimator remains bounded.

    \item \textbf{Simplified NELBO weights.}
    The LOOT estimator is weighted only by $\alpha_t'$, rather than $\alpha_t'/(1-\alpha_t)$.
    Under the log-linear schedule $\alpha_t=1-t$, the remaining weight is constant, $\alpha_t'=-1$, eliminating the endpoint-divergent weighting of the standard MDM estimator.
\end{enumerate}

\subsection{Connection to the Half-and-Half Loss}
\label{app:half-and-half}

A natural alternative to Rao--Blackwellization is to retain the masked-position estimator and add a symmetric supervision term for visible positions:
\begin{align}
\label{eq:loo-half}
\mathcal{L}_{\text{LOO-half}}(\theta)
=
\mathbb{E}_{\mathbf{x}}
\Bigg[
\int_0^1
\mathbb{E}_{M_t}
\Bigg(
&\frac{1}{2}
\frac{\alpha_t'}{1-\alpha_t}
\sum_{i\in M_t}
\log p_\theta(x_i \mid \mathbf{x}[M_t])
\nonumber\\
&+
\frac{1}{2}
\frac{\alpha_t'}{\alpha_t}
\sum_{i\notin M_t}
\log p_\theta
\left(x_i \mid \mathbf{x}[M_t\cup\{i\}]\right)
\Bigg)dt
\Bigg].
\end{align}

This is the half-and-half objective used by HollowDiff~\citep{zhao2024informed}. Concurrently, PGM~\citep{deschenaux2025partition} uses related complementary partition supervision with inverse-noise weighting, but predicts tokens in one partition from the opposite partition rather than satisfying the LOO property.

Like $\mathcal{L}_{\text{LOOT}}$, the half-and-half objective places loss on every position. Under the LOO property, it also equals the standard MDM NELBO in expectation. Each position is masked with probability $1-\alpha_t$, so marginalizing its own masking event gives an expected weight of
\[
\frac{1}{2}
\frac{\alpha_t'}{1-\alpha_t}
(1-\alpha_t)
+
\frac{1}{2}
\frac{\alpha_t'}{\alpha_t}
\alpha_t
=
\alpha_t'.
\]
Thus, applying the same per-position Rao--Blackwellization to the half-and-half estimator also yields the LOOT loss.

\paragraph{Numerical instability.}
Although the half-and-half and LOOT objectives agree in expectation, the former retains inverse-probability weights that diverge at both ends of the noise schedule. Writing $b_i=\mathbf{1}\{i\in M_t\}$, its per-position weight $w_i^{\mathrm{half}}=\frac{b_i}{2(1-\alpha_t)}+\frac{1-b_i}{2\alpha_t}$ has variance
\begin{equation}
\mathrm{Var}(w_i^{\mathrm{half}})
=\frac{(1-2\alpha_t)^2}{4\alpha_t(1-\alpha_t)}.
\end{equation}
By \eqref{eq:per-position-variance}, this contributes $\mathrm{Var}(w_i^{\mathrm{half}})\mathbb{E}[\ell_i^2]$ to each position's loss variance. Under the log-linear schedule $\alpha_t=1-t$, bounded log-probabilities and nonvanishing expected squared losses near low noise imply logarithmically divergent variance for the complete half-and-half estimator as the time-clipping threshold approaches zero; an analogous divergence arises near high noise under the corresponding nondegeneracy condition. In contrast, LOOT eliminates both inverse-probability weights and retains bounded estimator variance.

PGM~\citep{deschenaux2025partition} report occasional training-loss spikes under complementary masking, consistent with the numerical sensitivity of inverse-noise weighting. The LOOT loss therefore preserves the expected MDM NELBO while eliminating the endpoint-divergent inverse-probability weights of both the standard MDM and half-and-half estimators.

\section{Experimental details}
\label{app:training-details}

We follow the experimental protocol of~\citet{sahoo2024mdlm} closely, with
modifications as required by the LOOT architecture (Section~\ref{sec:method}).
Across both benchmarks, we optimize with Adam
(learning rate $3 \times 10^{-4}$, betas $(0.9, 0.999)$, $\epsilon = 10^{-8}$)
without weight decay, apply gradient clipping at $1.0$, and use a constant
learning rate schedule with $2{,}500$ steps of linear warmup. We maintain
an exponential moving average (EMA) of the weights with decay $0.9999$ and
use this EMA copy for evaluation. Models are trained in bf16 mixed
precision with a dropout rate of $0.1$. All runs use four NVIDIA B200 GPUs on a single node, with DDP across GPUs and gradient accumulation with a global batch size of $512$. Our
implementation builds on the official MDLM
codebase.\footnote{\url{https://github.com/kuleshov-group/mdlm}}

\subsection{Architectures}
\label{app:training-architectures}

\paragraph{MDM baseline.} The MDM baseline is a Diffusion Transformer
(DiT) with $12$ transformer blocks, $12$ attention heads, and a hidden
dimension of $768$.

\paragraph{LOOT.} LOOT uses the same per-block dimensions and depth as the
MDM baseline. It maintains three streams (forward $\vec{\mathbf{f}}$,
backward $\cev{\mathbf{b}}$, and output $\mathbf{o}$) at each layer
(\S\ref{sec:method-architecture}) under \emph{full} weight sharing: the three
streams share a single set of attention projections (QKV and output),
feedforward weights, and adaLN modulation parameters per block; the
forward and backward streams further share their pre-attention and
pre-MLP layer norms, while the output stream keeps its own layer norms.
The trainable parameter count therefore matches MDM. When initializing
from an MDM checkpoint, we copy each pretrained DiT block's QKV
projection, attention output projection, MLP, and adaLN modulation into
the shared parameters, and copy its two layer norms into both the
(shared) forward/backward layer norms and the output-stream layer norms.

\subsection{LM1B}
\label{app:training-lm1b}

For LM1B~\citep{lm1b2014}, we use the \texttt{bert-base-uncased} tokenizer
with a context length of $128$ tokens and apply sequence packing,
concatenating documents with the tokenizer's separator (\texttt{[SEP]}) as
the end-of-document marker. We train with a global batch size of $512$
for $1$M steps.

\subsection{OpenWebText}
\label{app:training-owt}

For OpenWebText (OWT)~\citep{Gokaslan2019OpenWeb}, we use the GPT-2
tokenizer with a context length of $1024$ tokens and apply sequence
packing, concatenating documents with the GPT-2 \texttt{<|endoftext|>}
token as the end-of-document marker. Since OWT does not provide an
official validation split, we reserve the last $100$K documents for
validation following~\citet{sahoo2024mdlm}. We train with a global batch
size of $512$ for $250$K steps (per-GPU micro-batch of $32$ with $4$
gradient accumulation steps).

\subsection{Self-Correction Policy Details}
\label{app:self-correction-policies}

All self-correction policies use the same \method{} primitive: given a
visible sequence \(x \in \mathcal{V}^N\), one forward pass returns
\(p_\theta(\cdot \mid x_{-i})\) for every position \(i\). The policies
differ only in how they choose editable positions \(A_r\) and replacements.

\begin{algorithm}[h]
\caption{\method{} Self-Correction}
\label{alg:loot-self-correction}
\begin{algorithmic}[1]
\Require Initial steps \(S_{\mathrm{init}}\), correction steps \(R\), policy \(\pi\)
\State Generate \(x^{(0)}\) with \(S_{\mathrm{init}}\) reverse denoising steps
\For{\(r=0,\ldots,R-1\)}
    \State Compute \(p_\theta(\cdot \mid x^{(r)}_{-i})\) for all \(i \in \{1,\ldots,N\}\)
    \State \(A_r, z_{A_r}, a_{A_r} \gets \pi(x^{(r)}, \{p_\theta(\cdot \mid x^{(r)}_{-i})\}_{i=1}^N)\)
    \For{\(i \in A_r\)}
        \If{\(a_i=1\)}
            \State \(x_i^{(r+1)} \gets z_i\)
        \Else
            \State \(x_i^{(r+1)} \gets x_i^{(r)}\)
        \EndIf
    \EndFor
    \State \(x_j^{(r+1)} \gets x_j^{(r)}\) for all \(j \notin A_r\)
\EndFor
\State \Return \(x^{(R)}\)
\end{algorithmic}
\end{algorithm}

In Algorithm~\ref{alg:loot-self-correction}, \(z_i\) is the proposed replacement token and \(a_i \in \{0,1\}\)
indicates whether the edit is accepted.

\paragraph{\method{}-Blocked-Gibbs.}
Partition positions into \(G\) groups \(B_1,\ldots,B_G\). At step \(r\),
choose $A_r = B_{1+(r \bmod G)}$.
For every \(i \in A_r\), sample
$z_i \sim p_\theta(\cdot \mid x^{(r)}_{-i})$, with
$a_i = 1$.
Thus all positions in the active block are resampled in parallel. In our experiments, we use \(G=4\) groups.

\paragraph{\method{}-Best-Improvement.}
For each position, compute
\[
\Delta_i =
\max_{v \neq x_i^{(r)}}
\log p_\theta(v \mid x^{(r)}_{-i})
-
\log p_\theta(x_i^{(r)} \mid x^{(r)}_{-i}) .
\]
Let \(A_r\) be the positions with largest positive \(\Delta_i\). For each
\(i \in A_r\), set
\[
z_i =
\arg\max_{v \neq x_i^{(r)}} p_\theta(v \mid x^{(r)}_{-i}),
\qquad a_i = 1 .
\]
This greedily applies the locally best token replacements.

\paragraph{\method{}-Metropolis-Hastings.}
For each selected position \(i \in A_r\), propose
$z_i \sim p_\theta(\cdot \mid x^{(r)}_{-i})$
and accept the replacement with probability
\[
\alpha_i =
\min\left(
1,
\frac{p_\theta(z_i \mid x^{(r)}_{-i})}
     {p_\theta(x_i^{(r)} \mid x^{(r)}_{-i})}
\right).
\]
Equivalently, draw \(u_i \sim \mathrm{Unif}(0,1)\) and set $a_i = \mathds{1}\{u_i \leq \alpha_i\}$.
\paragraph{\method{}-Blocked-Gibbs (guarded).}
Without nucleus truncation, high-temperature Gibbs updates occasionally insert low-probability tokens that corrupt the context of neighboring positions, and the chain can drift into repetitive fixed points (Appendix~\ref{app:topp1}). The guarded variant uses the same blocks and proposals as blocked Gibbs but accepts a draw only if it does not lower the leave-one-out likelihood of its position,
\[
a_i = \mathds{1}\{p_\theta(z_i \mid x^{(r)}_{-i}) \geq p_\theta(x_i^{(r)} \mid x^{(r)}_{-i})\}.
\]
This deterministic acceptance rule prevents the collapse at every budget and temperature we tested, at a modest cost in generative perplexity where plain blocked Gibbs is stable. We use it only in the experiments without nucleus truncation. All policies use only the trained \method{} model.

\subsection{Sampling evaluation}
\label{app:sampling-eval-details}

We evaluate unconditional generation on OpenWebText at sequence length
$N=1024$. Each frontier point uses 512 independently generated samples.
All samplers use nucleus sampling with $p=0.9$, sweeping temperatures
$0.7 \leq \tau \leq 1.1$.

We measure sampling compute in number of forward evaluations (NFE).
For MDM, ReMDM, and the uncorrected \method{} baseline, NFE counts
denoising steps. For \method{} with self-correction,
$\mathrm{NFE}=S_{\mathrm{init}}+R$, where $S_{\mathrm{init}}$ and $R$
denote initial denoising and refinement steps, respectively.
Unless otherwise stated, we allocate half the budget to each:
$S_{\mathrm{init}}=R=\mathrm{NFE}/2$.
Correction policies are detailed in Appendix~\ref{app:self-correction-policies}.

We evaluate generation quality using generative perplexity (GenPPL),
computed as the exponential of the mean negative log-likelihood
under GPT-2 Large~\citep{radford2019language}, with teacher forcing
after decoding and retokenizing generated samples. We measure diversity
using sample entropy $\widetilde H(q)$, defined as the average empirical
unigram entropy of each generated sequence's token IDs, before
detokenization. Lower GenPPL indicates higher generation quality,
while higher entropy indicates greater token diversity. We report
quality--diversity frontiers across sampling temperatures.

\subsection{Sampling without nucleus truncation}
\label{app:topp1}

\begin{figure*}[h]
    \centering
    \begin{subfigure}[h]{0.58\textwidth}
        \centering
        \includegraphics[width=\linewidth]{
            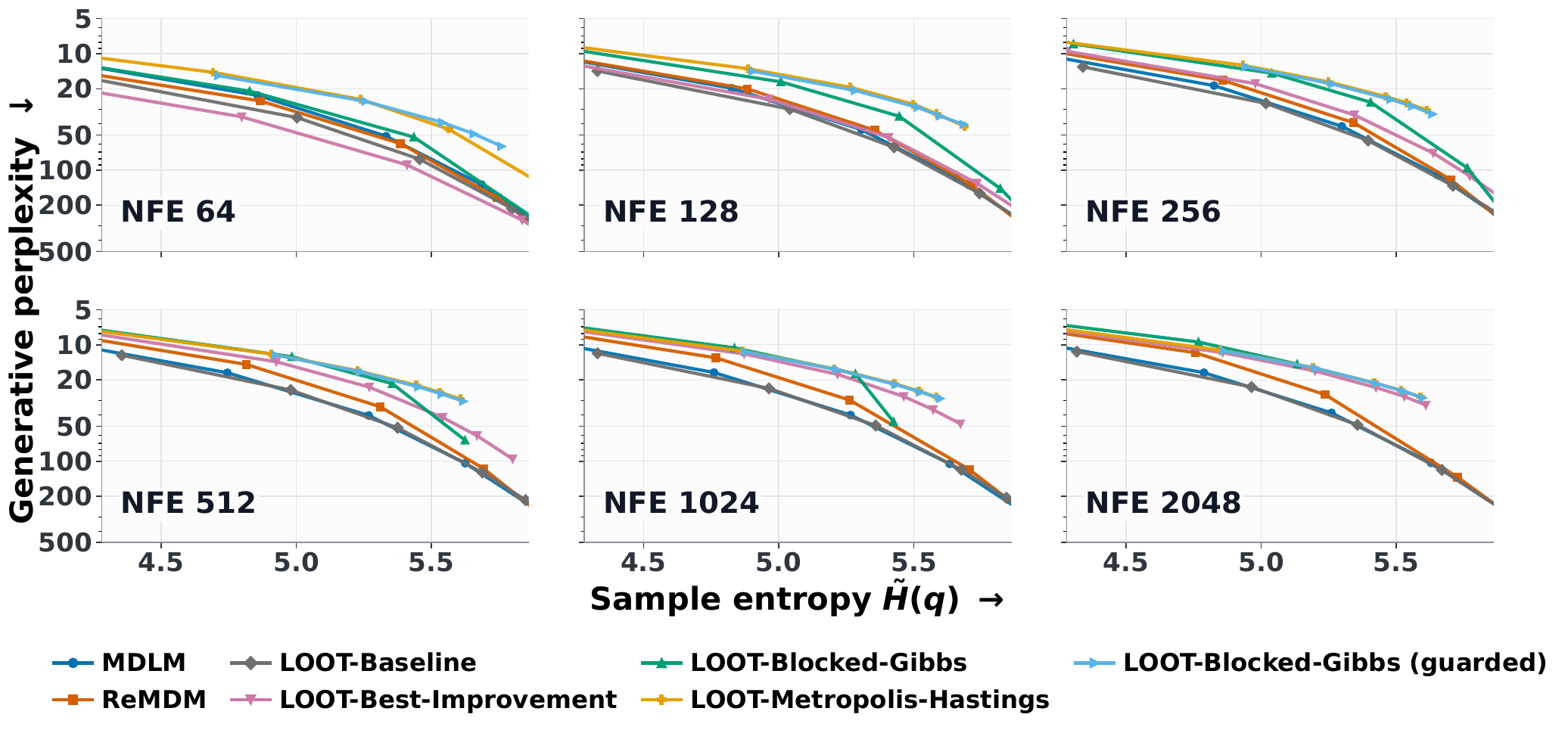
        }
        \caption{Quality--diversity frontiers across NFE budgets.}
        \label{fig:frontiers-topp1}
    \end{subfigure}
    \hfill
    \begin{subfigure}[h]{0.4\textwidth}
        \centering
        \includegraphics[width=\linewidth]{
            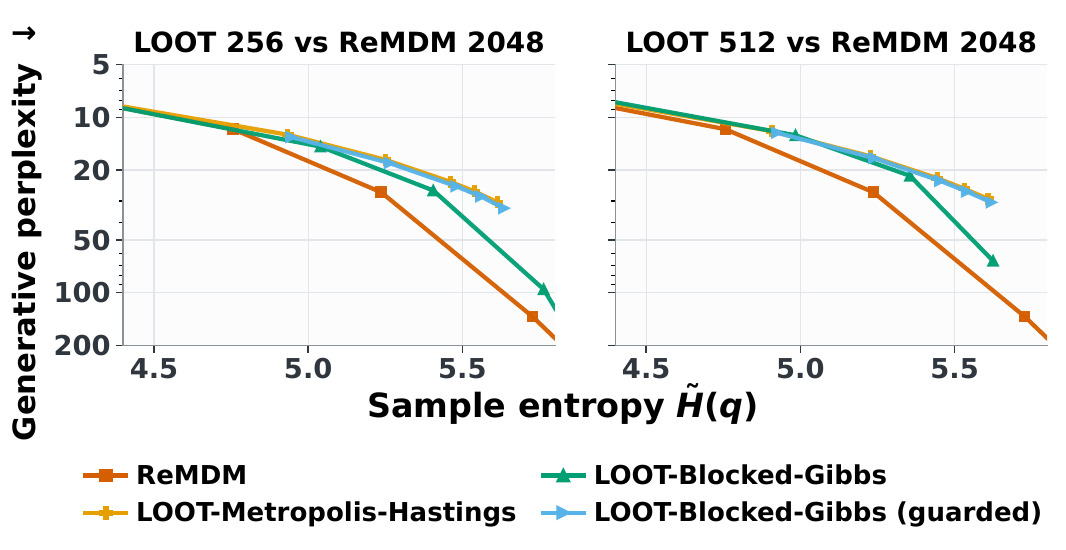
        }
        \caption{Unequal-compute comparison.}
        \label{fig:compute-topp1}
    \end{subfigure}

    \caption{\textbf{Generation quality without nucleus truncation
    (top-$p=1$).}
    (a) Quality--diversity frontiers using the protocol of
    Figure~\ref{fig:sampling-frontier-all-nfe}, with a logarithmic
    perplexity axis. Temperature sweeps are truncated where
    entropy stops increasing; guarded Blocked Gibbs remains
    stable across budgets.
    (b) \method{} self-correction at 256 and 512 NFEs versus
    ReMDM at 2048 NFEs, as in
    Figure~\ref{fig:sampling-compute-summary}(a).}
    \label{fig:topp1-results}
\end{figure*}

Figures~\ref{fig:frontiers-topp1} and~\ref{fig:compute-topp1} repeat the comparison of Section~\ref{sec:sampling-finetuning} with top-$p = 1$, i.e., sampling from the full categorical distribution at every step. Without truncation, MDLM and ReMDM exceed a generative perplexity of $100$ at temperature $1.0$ and above, so the perplexity axis is logarithmic. Plain blocked Gibbs collapses into repetition at temperature $1.0$ and above: high-temperature draws occasionally insert low-probability tokens, the corrupted context then favors copying neighboring tokens, and the chain drifts into a low-entropy fixed point. We cut each sweep where entropy stops increasing with temperature. The guarded variant of Appendix~\ref{app:self-correction-policies} does not collapse at any budget, and together with the Metropolis-style corrector it dominates ReMDM at 2048 NFE from 128 NFE onward over the whole entropy range.

\subsection{Full MAUVE results}
\label{app:mauve-full}

Table~\ref{tab:mauve5k} reports MAUVE for every sampler at every NFE budget, on 5000 samples per setting, together with generative perplexity and entropy on the same samples. We follow the MAUVE evaluation protocol of \citet{wang2025remasking}, ensuring direct comparability with their reported results. Blocked Gibbs achieves the highest MAUVE at every budget from 128 NFE onward.

\begin{table}[h!]
\centering
\caption{\textbf{Full sampling results on OpenWebText.}
Generative perplexity, sample entropy, and MAUVE across NFE
budgets, with top-$p=0.9$ and the indicated sampling temperature
$T$. Best genPPL and MAUVE values at each NFE are bolded.
BG, BI, and MH denote Blocked Gibbs, Best Improvement, and
Metropolis--Hastings, respectively.}
\label{tab:mauve5k}
\footnotesize
\setlength{\tabcolsep}{3pt}
\renewcommand{\arraystretch}{1.05}

\begin{minipage}[t]{0.46\textwidth}
\centering
\resizebox{\linewidth}{!}{%
\begin{tabular}{@{}rlcrrr@{}}
\toprule
\textbf{NFE} & \textbf{Method} & $T$
& \textbf{genPPL}
& \textbf{Entropy}
& \textbf{MAUVE} \\
\midrule
64 & MDLM & 1 & 50.4 & 5.42 & 0.011 \\
 & ReMDM & 1 & 57.8 & 5.47 & 0.009 \\
 & \method{}-Baseline & 0.95 & 58.7 & 5.42 & 0.007 \\
\rowcolor{blue!10}
 & \method{}-BG & 1 & 47.8 & 5.56 & 0.015 \\
 & \method{}-BI & 0.95 & 71.2 & 5.42 & 0.006 \\
 & \method{}-MH & 1 & \textbf{29.7} & 5.43 & \textbf{0.017} \\
\midrule
128 & MDLM & 1.05 & 68.0 & 5.56 & 0.010 \\
 & ReMDM & 1 & 41.6 & 5.41 & 0.035 \\
 & \method{}-Baseline & 1 & 66.1 & 5.52 & 0.008 \\
\rowcolor{blue!10}
 & \method{}-BG & 1 & 27.9 & 5.48 & \textbf{0.118} \\
 & \method{}-BI & 1 & 59.8 & 5.54 & 0.010 \\
 & \method{}-MH & 1.05 & \textbf{24.7} & 5.48 & 0.095 \\
\midrule
256 & MDLM & 1.05 & 63.1 & 5.53 & 0.014 \\
 & ReMDM & 1.05 & 53.9 & 5.54 & 0.040 \\
 & \method{}-Baseline & 1 & 55.4 & 5.47 & 0.010 \\
\rowcolor{blue!10}
 & \method{}-BG & 1 & \textbf{20.1} & 5.40 & \textbf{0.408} \\
 & \method{}-BI & 1 & 36.3 & 5.44 & 0.035 \\
 & \method{}-MH & 1.05 & 21.2 & 5.43 & 0.312 \\
\bottomrule
\end{tabular}%
}
\end{minipage}%
\hspace{0.06\textwidth}%
\begin{minipage}[t]{0.46\textwidth}
\centering
\resizebox{\linewidth}{!}{%
\begin{tabular}{@{}rlcrrr@{}}
\toprule
\textbf{NFE} & \textbf{Method} & $T$
& \textbf{genPPL}
& \textbf{Entropy}
& \textbf{MAUVE} \\
\midrule
512 & MDLM & 1.05 & 60.8 & 5.52 & 0.015 \\
 & ReMDM & 1.05 & 44.9 & 5.50 & 0.091 \\
 & \method{}-Baseline & 1 & 50.8 & 5.45 & 0.010 \\
\rowcolor{blue!10}
 & \method{}-BG & 1.05 & \textbf{20.0} & 5.45 & \textbf{0.549} \\
 & \method{}-BI & 1.05 & 32.3 & 5.48 & 0.072 \\
 & \method{}-MH & 1.1 & 22.8 & 5.49 & 0.366 \\
\midrule
1024 & MDLM & 1.05 & 59.4 & 5.51 & 0.017 \\
 & ReMDM & 1.05 & 37.6 & 5.45 & 0.242 \\
 & \method{}-Baseline & 1 & 48.8 & 5.43 & 0.012 \\
\rowcolor{blue!10}
 & \method{}-BG & 1.1 & \textbf{21.1} & 5.51 & \textbf{0.580} \\
 & \method{}-BI & 1.1 & 29.2 & 5.52 & 0.201 \\
 & \method{}-MH & 1.1 & 22.1 & 5.47 & 0.432 \\
\midrule
2048 & MDLM & 1.05 & 58.5 & 5.50 & 0.018 \\
 & ReMDM & 1.05 & 31.2 & 5.40 & 0.348 \\
 & \method{}-Baseline & 1 & 47.6 & 5.42 & 0.012 \\
\rowcolor{blue!10}
 & \method{}-BG & 1.1 & \textbf{17.5} & 5.46 & \textbf{0.602} \\
 & \method{}-BI & 1.1 & 24.3 & 5.48 & 0.401 \\
 & \method{}-MH & 1.1 & 21.8 & 5.47 & 0.478 \\
\bottomrule
\end{tabular}%
}
\end{minipage}

\end{table}

\subsection{Ablations}
\label{app:ablations}

This section presents two ablations complementing the language modeling results in Section~\ref{sec:val-ppl-results}. First, we disentangle the effects of architecture and training objective across sequence lengths on OpenWebText (Table~\ref{tab:length_ablation}), investigating the likelihood gap between \method{} and standard MDMs at longer sequences. Second, we evaluate the effects of merging frequency and training objective on LM1B (Table~\ref{tab:hollow_ablation}), demonstrating how both improvements contribute to the performance of \method{}.

\begin{table*}[h!]
    \centering

    \begin{minipage}[h!]{0.40\textwidth}
        \centering
        \captionof{table}{\textbf{Sequence length ablations.}
        Validation perplexity ($\downarrow$) on OpenWebText at
        $N=128$ (1M training steps) and $N=1024$ (250K training steps),
        with batch size 512. We compare standard MDLM against
        the \method{} architecture trained with either the
        standard MDLM loss or our \method{} objective.}
        \label{tab:length_ablation}

        \vspace{0.5em}
        {\footnotesize
        \setlength{\tabcolsep}{3pt}
        \begin{tabular}{lcc}
        \toprule
        \textbf{Model} & $N=128$ & $N=1024$ \\
        \midrule
        MDLM (DiT) & 35.23 & \textbf{25.19} \\
        \method{} + MDLM loss & 37.40 & 29.63 \\
        \rowcolor{blue!10}
        \method{} + \method{} loss & \textbf{34.51} & 26.83 \\
        \bottomrule
        \end{tabular}
        }
    \end{minipage}
    \hfill
    \begin{minipage}[h!]{0.58\textwidth}
        \centering
        \captionof{table}{\textbf{Hollow transformer ablations on LM1B.}
        Validation perplexity ($\downarrow$) at $N=128$,
        batch size 256, after 200K training steps.
        Hollow variants use the \method{} NELBO unless stated.
        $^{*}$\method{} with merges every $m$ layers, following
        HollowDiff~\citep{zhao2024informed}, isolating merging frequency.
        }
        \label{tab:hollow_ablation}

        \vspace{0.5em}
        {\footnotesize
        \setlength{\tabcolsep}{4pt}
        \begin{tabular}{llc}
        \toprule
        \textbf{Model} & \textbf{Merge} & \textbf{PPL ($\downarrow$)} \\
        \midrule
        DiT + MDLM loss & n/a & 39.80 \\
        \midrule
        Hollow transformer~\citep{sun2022score}
            & final layer & 41.07 \\
        HollowDiff$^{*}$ ($m=3$)
            & every 3 layers & 37.91 \\
        HollowDiff$^{*}$ ($m=2$)
            & every 2 layers & 37.23 \\
        \rowcolor{blue!10}
        \method{} ($m=1$)
            & every layer & \textbf{36.23} \\
        \method{} + half-and-half loss
            & every layer & 39.35 \\
        \bottomrule
        \end{tabular}
        }
    \end{minipage}

\end{table*}

\subsection{Self-correction with the MDLM loss}
\label{app:hybrid}

To disentangle the benefits of self-correction from dense supervision, we evaluate the \method{} architecture trained with the standard masked-only MDLM loss. We use the $N=1024$ OWT model from Table~\ref{tab:length_ablation}, selecting its best-validation checkpoint after 250k steps, trained with the standard MDM loss. Although visible tokens receive no direct supervision, the architecture still provides leave-one-out conditionals for self-correction. Table~\ref{tab:hybrid} compares ancestral sampling with Blocked Gibbs correction using the same model weights. Self-correction reduces generative perplexity by $2.4\times$--$4.3\times$ across NFE budgets. At 256 NFEs, Blocked Gibbs reduces genPPL from $62.3$ to $19.9$, matching the corrected model trained with our all-position objective. Even at 2048 NFEs, it achieves genPPL $12.9$, compared with $19.0$ for ReMDM and $37.4$ for MDLM. These results demonstrate that effective self-correction is enabled by the leave-one-out architecture, even without dense supervision, while our objective independently improves likelihood and uncorrected generation quality.

\begin{table*}[h]
    \centering
    \footnotesize
    \caption{\textbf{Self-correction with the standard MDM loss.}
    Generative perplexity ($\downarrow$) and sample entropy on
    OpenWebText ($N=1024$), using $T=1.0$, top-$p=0.9$,
    and 512 samples per setting. Ancestral sampling uses
    standard MDM unmasking without correction; Blocked Gibbs
    allocates half of the NFEs to initial sampling and half
    to refinement ($S_\text{init}=R=\mathrm{NFE}/2$).}
    \label{tab:hybrid}

    \setlength{\tabcolsep}{3pt}
    \resizebox{0.8\textwidth}{!}{%
    \begin{tabular}{llcccccc|cccccc}
    \toprule
    & & \multicolumn{6}{c|}{Gen. PPL ($\downarrow$)}
        & \multicolumn{6}{c}{Entropy} \\
    \midrule
    \rowcolor{lightgray!25}
    \multicolumn{2}{r}{$\mathrm{NFE} =$}
    & 64 & 128 & 256 & 512 & 1024 & 2048
    & 64 & 128 & 256 & 512 & 1024 & 2048 \\

    \multicolumn{2}{l}{MDLM}
    & 49.9 & 43.2 & 39.6 & 37.8 & 37.6 & 37.4
    & 5.42 & 5.37 & 5.35 & 5.32 & 5.34 & 5.33 \\

    \multicolumn{2}{l}{ReMDM}
    & 56.8 & 40.6 & 33.1 & 27.3 & 22.4 & 19.0
    & 5.46 & 5.40 & 5.36 & 5.31 & 5.24 & 5.20 \\

    \midrule
    \method{} + MDM loss & Baseline
    & 90.6 & 70.6 & 62.3 & 58.2 & 56.9 & 55.6
    & 5.55 & 5.51 & 5.49 & 5.47 & 5.46 & 5.45 \\

    \rowcolor{blue!10}
    & Blocked Gibbs
    & 38.1 & 25.9 & 19.9 & 16.7 & 14.5 & 12.9
    & 5.53 & 5.44 & 5.36 & 5.32 & 5.26 & 5.21 \\

    \midrule
    \method{} + \method{} loss & Baseline
    & 91.1 & 65.4 & 55.0 & 49.5 & 47.5 & 47.5
    & 5.59 & 5.52 & 5.47 & 5.45 & 5.42 & 5.42 \\

    & Blocked Gibbs
    & 47.6 & 28.3 & 19.9 & 16.2 & 14.0 & 12.4
    & 5.56 & 5.48 & 5.40 & 5.33 & 5.29 & 5.24 \\

    \bottomrule
    \end{tabular}%
    }
\end{table*}

\subsection{Additional sampling experiments}

\begin{figure*}[h!]
    \centering
    \begin{subfigure}{0.49\textwidth}
        \centering
        \includegraphics[width=\linewidth]{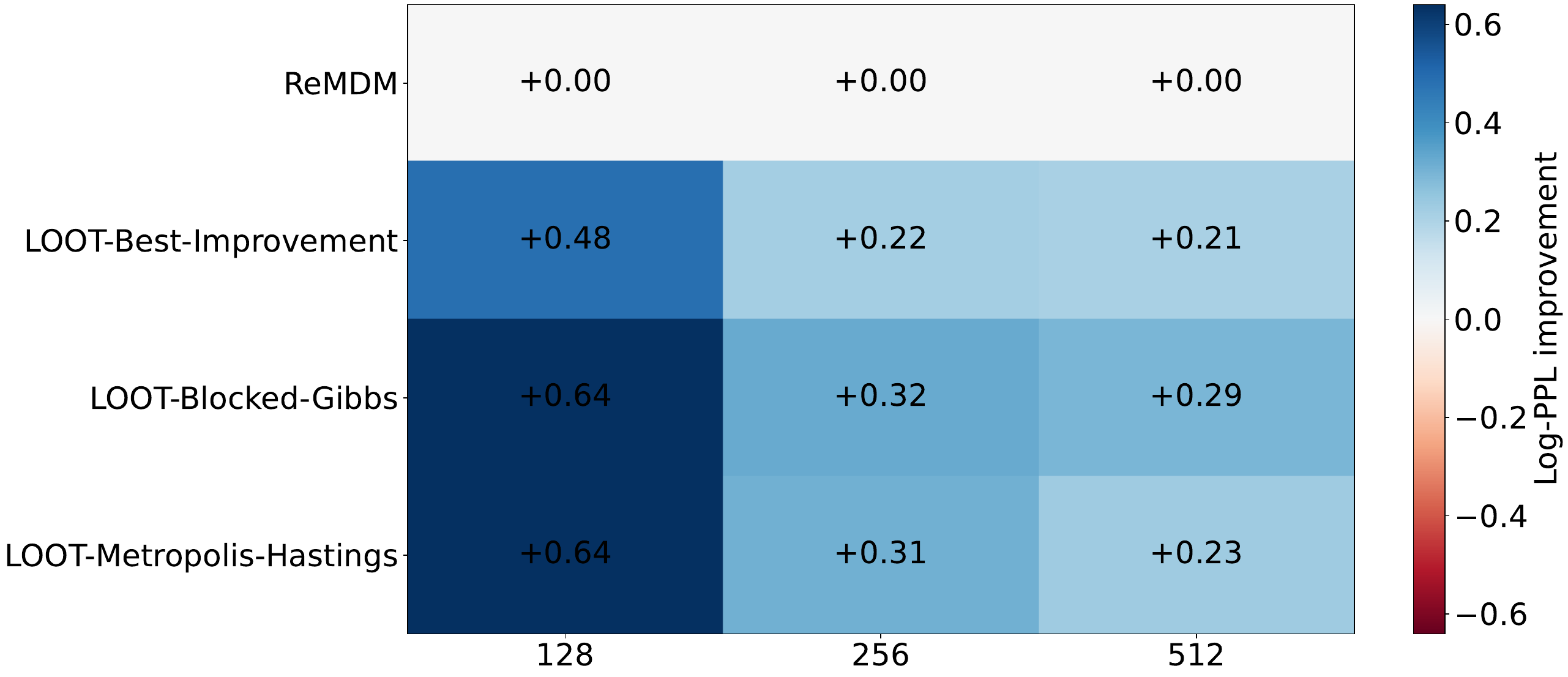}
        \caption{Best improvement in $4.5 \leq \widetilde{H}(q) \leq 5.5$.}
    \end{subfigure}
    \hfill
    \begin{subfigure}{0.49\textwidth}
        \centering
        \includegraphics[width=\linewidth]{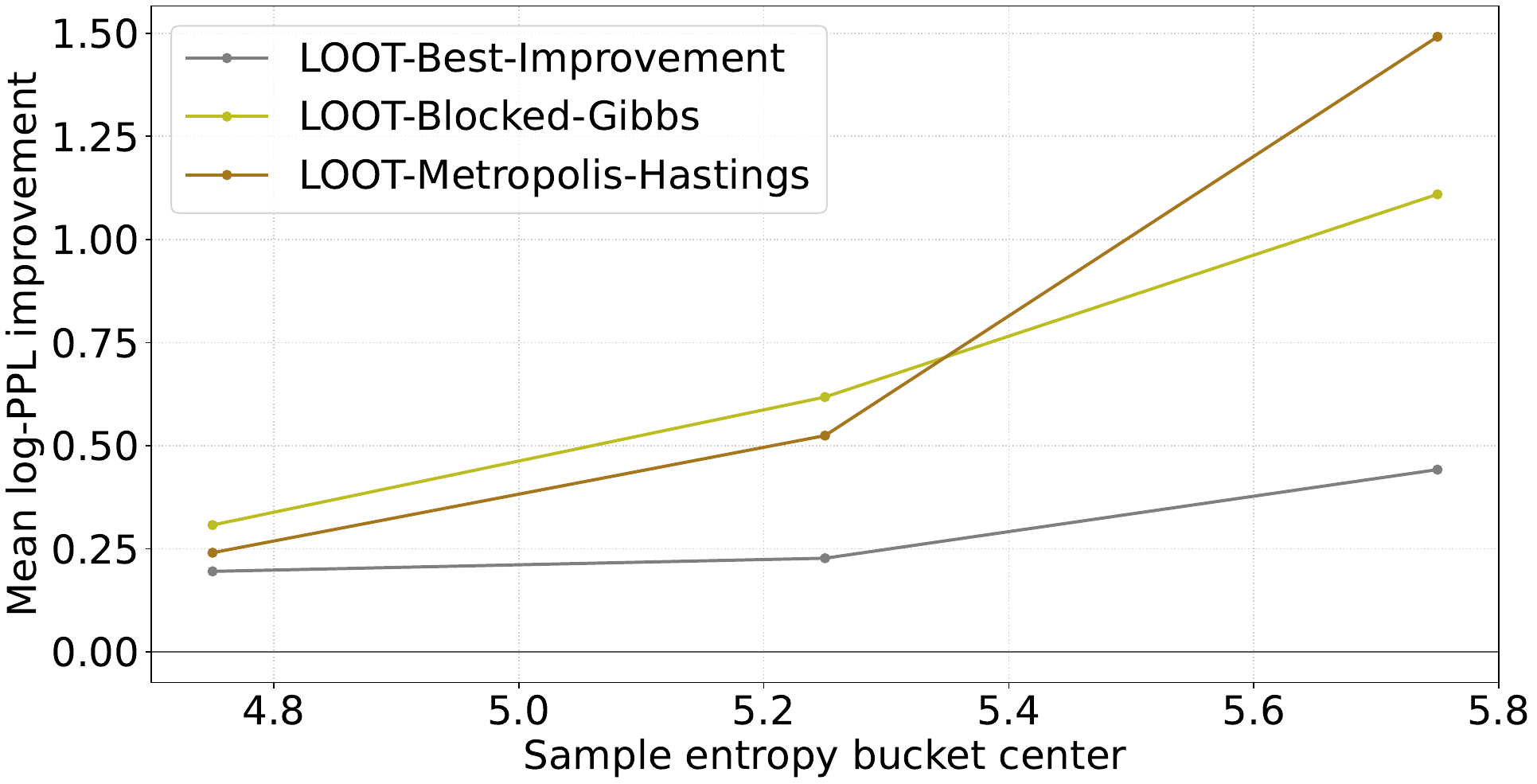}
        \caption{Mean improvement by entropy bucket.}
    \end{subfigure}
    \caption{\textbf{\method{} gains are robust across entropy and compute.}
    We report log-PPL improvement over ReMDM, where positive values indicate lower generative perplexity. Improvements persist across NFE budgets and entropy buckets.}
    \label{fig:sampling-robustness}
\end{figure*}

\paragraph{Robustness.} Figure~\ref{fig:sampling-robustness} shows that the sampling gains are not isolated to a single temperature or policy. Across entropy buckets, all \method{} correction policies improve log-PPL over ReMDM, with the largest gains in the high-entropy regime where ReMDM degrades most sharply.

\begin{figure*}[h!]
    \centering
    \includegraphics[width=0.5\textwidth]{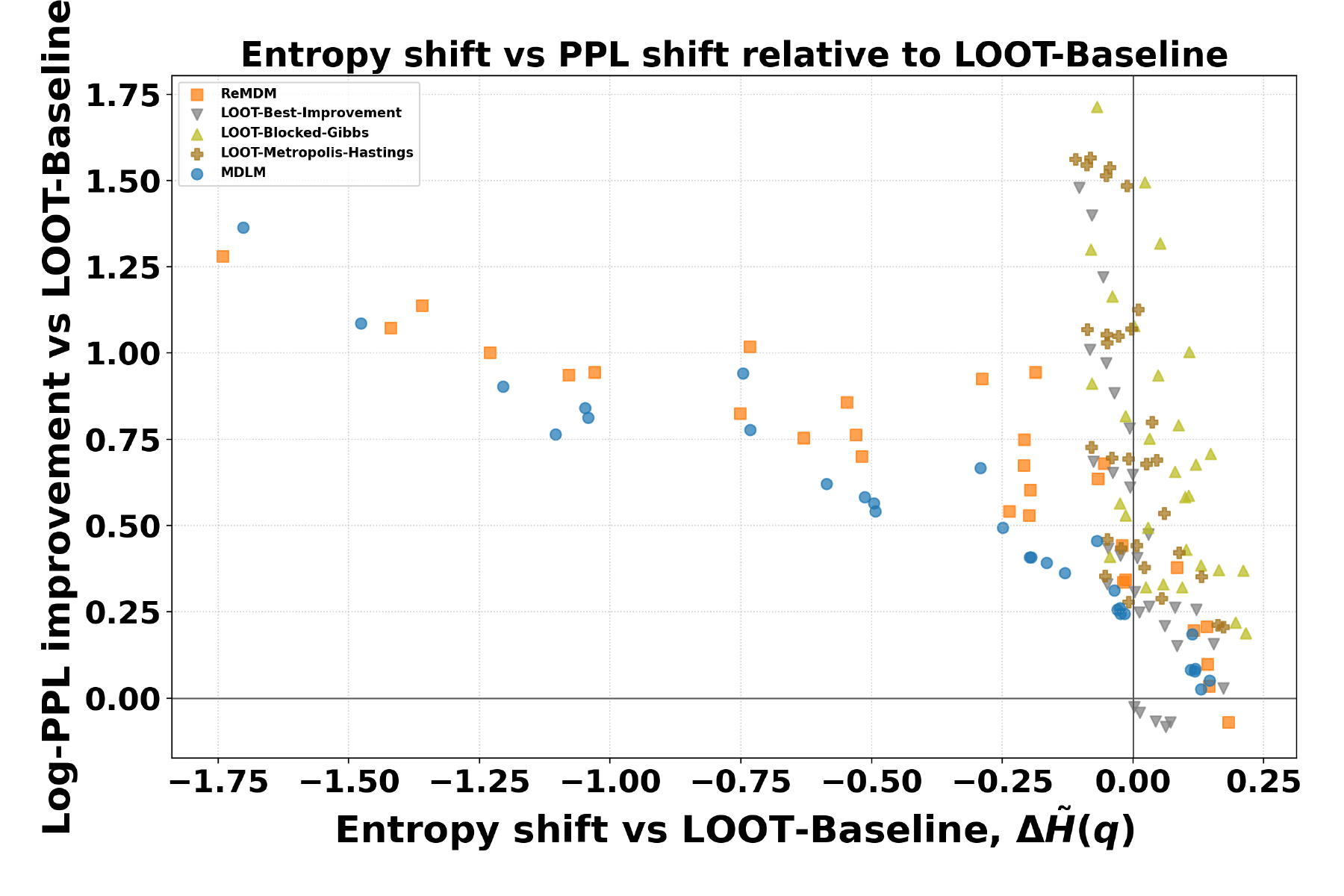}
    \caption{\textbf{\method{} improves quality without collapsing entropy.}
    Each point is matched to the uncorrected \method{} baseline at the same temperature and NFE. Positive vertical values indicate lower log-PPL than the baseline; horizontal values near zero indicate little entropy loss. \method{} correction policies concentrate in the high-improvement, low-entropy-shift region.}
    \label{fig:sampling-entropy-shift}
\end{figure*}

\begin{figure*}[t]
    \centering
    \includegraphics[width=0.8\textwidth]{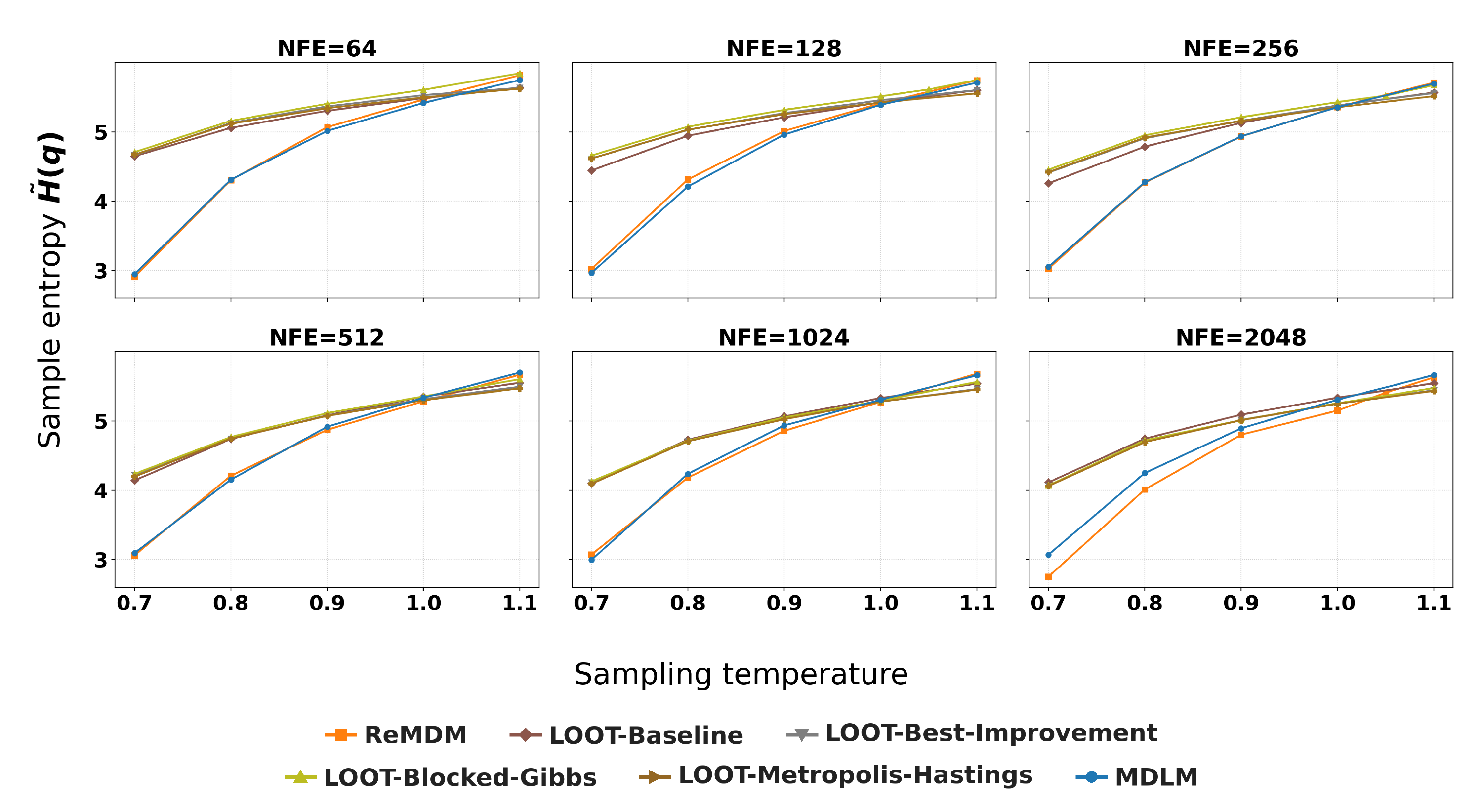}
    \caption{\textbf{Sample entropy response across NFE budgets.}}
    \label{fig:sampling-temperature-entropy}
\end{figure*}

\paragraph{Temperature response.}
Figure~\ref{fig:sampling-temperature-entropy} and Figure~\ref{fig:sampling-entropy-shift} show that \method{} self-correction maintains higher entropy than MDLM and ReMDM baselines, across NFE budgets.

\newpage
\subsection{Qualitative OpenWebText generations}
\label{app:qualitative-owt-samples}

We present qualitative generations from models trained on OpenWebText. All samples use length 1024, temperature 1.0, top-$p=0.9$,
and 1024 model evaluations.

\paragraph{LOOT-Blocked-Gibbs, $G=4$.}
\begin{quote}\small
The Bitcoin community is moving strongly toward altcoins right now, with the
latest altcoin being Monero. If you ask me, I say the most important thing to
know about mining and whether altcoins are about money. One of the most
exciting aspects of mining is getting an overview of what is going on in the
Bitcoin ecosystem and getting to know everything going on along the way. The
Ultimate Guide to Bitcoin Mining tells you everything that you really need to
know about the basics of mining in the Bitcoin ecosystem. ``Bitcoin.com's Mining
Guide is a comprehensive and comprehensive guide that covers all the basics of
mining.'' Want more? Do not forget to join the Bitcoin Forum mailing list, where
we can answer your questions.
\end{quote}

\begin{quote}\small
One of our latest and greatest 3D printing products is 3D printing. Our new ASUS
3D printer is innovative. It is not made straight out of cardboard or paper but
it is made with a 3D printing system that is designed for delivering high
quality products. So that sounds simple, but while we are trying to accomplish
things like this, we have much more work to do and much more time than is
needed to come up with a high quality product. We also have a strong commitment
and commitment to delivering quality products to people who want to be our
customers. To deliver quality products and services, we need to make sure
customer satisfaction is our number one priority.
\end{quote}

\begin{quote}\small
``It has made such a difference in my life,'' Mo said. ``I now have a better
life and a better life than I ever had. We are a family. We have always wanted
our dad to come back to Scotland.'' Mo, 28, still hopes he will find work. Ever
since he was released, he has given up looking for work and has been out of work
for years. But now he wants to take care of himself a bit more. Now that one of
Mo's neighbours has come knocking on his door, he has two rental properties
lying empty in a desperate attempt to find a job.
\end{quote}

\paragraph{LOOT-Best-Improvement.}
\begin{quote}\small
The biggest challenge for us is we have to improve working conditions and the
quality of our work,'' he said. ``The problem is the fact that when you have the
backing of the big companies, the same thing is creating a much bigger and
riskier challenge to working conditions.'' World Economic Outlook 2016 The
United States and China receive 40\% of India's investment and trade-related
return on investment, while Saudi Arabia, Chinese and Indian companies are also
receiving higher returns than the US. Last year, the OECD reported that 60\% of
US companies, and 40\% were as big as the UK and Germany. The US and China are
the only two countries where the past year has brought bad news.
\end{quote}

\begin{quote}\small
The video footage from the Clifton Ballet show was released Wednesday, after
Greenpeace and the Wilberforce Foundation filed a lawsuit in support of the
pesticide ban. ``This is the largest smidgen of pesticide use that has ever had
an adverse effect on health, and the court of appeals has allowed a large number
of chemicals that cause cancer to be splashed into soil and water,'' Greenpeace
said. ``The debate over pesticides also sends a message that they should be
banned in order to reduce the harm caused by them, such as air pollution, and
other harmful bioactive substances,'' Greenpeace said in a statement.
\end{quote}

\begin{quote}\small
McGuidley's report on stock prices in late August, just a week after he
resigned, said he is convinced there is evidence that many traders and service
providers believe that a market correction will occur, and that even very high
stock prices will benefit from it. ``BnB Capital Markets probably would have had
a similar effect if the trading volume fell and prices were rising seasonally,''
he said. Black Friday, after all, rose 5 percent in August, and posted a few big
gains in stock prices since then.
\end{quote}

\paragraph{LOOT-Metropolis-Hastings.}
\begin{quote}\small
In the area where the park was last shut down in 2010, the Forest Service has
scrubbed the fallen trees from the park's Grote and other mulch trees, a
property donated to President Obama by climate-change activists. That has led to
increased logging, for which the Forest Service continues to collect data every
year. ``The U.S. Forest Service has been able to collect that data and know that
on the basis of that data, many trees were lost in the park,'' said Michael
Moriarty, of the National Resource Center for Rainforest. ``The data for every
tree in the park is collected and compared with data from previous years. Many
of those are trees that were lost that have not been collected by the federal
Forest Service yet.''
\end{quote}

\begin{quote}\small
It is the latest development in a year where BlackBerry has found it nearly
impossible to roll out a smartphone globally and to expand its reach beyond the
United States. The company plans to launch about one-hundred new smartphones
around the world, with most of them being expected to roll out sooner than the
batch of more than 1,000 that are expected to launch. The company says it will
be the first BlackBerry in the world and the first smartphone to be officially
launched this year.
\end{quote}

\begin{quote}\small
Disney, which owns Pixar's stake in the film's parent company. How big is the
Disney deal for Pixar? Disney's estimates of the total outlay were incorrect as
of Wednesday morning, per the company's website. An earlier report said that
Disney bought Disney Animation from Pixar after Pixar acquired it for about
\$100 million. The deal allows Pixar to be part of Disney for five years. The
Disney deal is one of the reasons the company has had to undergo a major
restructuring because of the reorganization.
\end{quote}

\paragraph{ReMDM.}
\begin{quote}\small
Opponents say it runs the huge risk of market consolidation, which puts
broadband and other telecoms-based services at risk. Meanwhile, wireless
technology deployment costs more than US\$12bn and requires major investment in
innovation. Asked about Verizon's proposal, Wheeler fired back, saying the
company's proposal would leave millions of customers with overpriced services.
``It is driving up the prices of providers, especially for the biggest providers
to consumers,'' he said. ``That is what we have created.'' Verizon declined to
comment Thursday. Wheeler's proposal is set for next year's hearings on telecom,
where the other major carriers, such as Verizon and Charter, are being
hammered. Verizon's proposal would open up the possibility of an overhaul of the
Code of Federal Regulations, which governs the wireless industry.
\end{quote}

\begin{quote}\small
Ian Desmond could have made the team in the 2015 playoffs and suited up for
plate duty. Instead, he has received some much needed help in the walk-out
department. ``I was always trying to make the contact and take advantage of the
movement ability of opposing hitters,'' Anderson said after his Feb. 18 loss to
the White Sox over the Nationals. ``I did not do that at the plate because I
felt like I had to. Now, there is the thing about being on the field as the
hitter and getting the walks.''
\end{quote}

\begin{quote}\small
Since then, George Martin and I have done some great stuff, and we were both
able to create the world and get to be a part of the story. Were there
influences in the casts and sets? Did you meet in one scene? Did you meet any of
the different creatives? We met a lot of fans. I am really excited about what
George has done and what I have written. There is a fascinating book that I
think can be read in its own way. How fun is it still exploring that little
background that you know about Game of Thrones?
\end{quote}

\section{Checkpoint Compatibility and Hybrid Modeling}
\label{app:compatibility}

\paragraph{Recovering bidirectional attention.}
An interesting consequence of \method{}'s full parameter sharing is that the architecture can recover standard bidirectional MDM computation without introducing additional parameters. Recall that \method{} maintains three streams: a forward stream $\vec{\mathbf{f}}$, a backward stream $\cev{\mathbf{b}}$, and an output stream $\mathbf{o}$. To recover bidirectional attention, we copy the output stream into both directional streams before every layer:
\begin{equation}
    \vec{\mathbf{f}}^{(\ell)} \leftarrow \mathbf{o}^{(\ell)},
    \qquad
    \cev{\mathbf{b}}^{(\ell)} \leftarrow \mathbf{o}^{(\ell)}.
\end{equation}
We additionally allow the output stream to attend to its own position during merging. Since all three streams now share the same representations and attention parameters, their contributions can be combined to recover standard bidirectional self-attention exactly.

To see this, let $q_i$, $k_j$, and $v_j$ denote the query, key, and value projections of the shared representations, and define the attention scores $s_{ij}=q_i^\top k_j/\sqrt{d}$. We partition the attention computation at position $i$ into left, right, and self contributions, with normalization factors
\begin{equation}
\begin{aligned}
    Z_i^{(\mathrm{left})} &= \sum_{j<i} e^{s_{ij}}, \\
    Z_i^{(\mathrm{right})} &= \sum_{j>i} e^{s_{ij}}, \\
    Z_i^{(\mathrm{self})} &= e^{s_{ii}}.
\end{aligned}
\end{equation}
Let $u_i^{(\mathrm{left})}$ and $u_i^{(\mathrm{right})}$ denote the softmax-normalized attention outputs over the respective contexts. The merged output, including self-attention, is
\begin{equation}
\begin{aligned}
    \tilde{o}_i
    &=
    \frac{
        Z_i^{(\mathrm{left})}u_i^{(\mathrm{left})}
        + Z_i^{(\mathrm{self})}v_i
        + Z_i^{(\mathrm{right})}u_i^{(\mathrm{right})}
    }{
        Z_i^{(\mathrm{left})}
        + Z_i^{(\mathrm{self})}
        + Z_i^{(\mathrm{right})}
    }\\
    &=
    \frac{\sum_{j=1}^{N} e^{s_{ij}}v_j}
         {\sum_{j=1}^{N} e^{s_{ij}}}\\
    &= \operatorname{Attn}(q_i,K,V).
\end{aligned}
\end{equation}
Thus, the three-way merge is mathematically identical to bidirectional softmax attention. With shared parameters, consistent positional encodings, and matching attention masks, applying this construction at every layer recovers a standard bidirectional transformer. In particular, the directional-stream computations can be omitted entirely in this mode, reducing \method{} to the computational cost of a standard MDM forward pass.

\paragraph{Self-attention in the output stream.}
We also explored incorporating the self contribution into the output stream during standard leave-one-out training. This preserves the LOO property because the output representation $\mathbf{o}_i$ is already independent of the input token $x_i$. However, we found no consistent improvement in language modeling performance, with some configurations performing slightly worse. Furthermore, combining self-attention with our efficient sliding-window merging mechanism complicates backpropagation through the attention normalization, requiring FlexAttention or a custom differentiable implementation. We therefore omit this contribution from our default \method{} architecture.

\paragraph{Hybrid training and inference.}
The equivalence to bidirectional attention suggests a hybrid approach that combines the computational efficiency of standard MDMs with the dense supervision and self-correction capabilities of \method{}. At high masking rates, standard MDM training already supervises most positions, so the additional benefit of leave-one-out prediction may be limited despite its approximately $2.2\times$ forward-pass FLOP overhead. As more tokens become visible, however, leave-one-out prediction becomes increasingly valuable, providing supervision at previously unmasked positions and enabling correction of earlier predictions.

A promising direction is therefore to train a single model in both modes, using standard bidirectional MDM computation at high masking rates and switching to leave-one-out computation at lower masking rates. The same strategy could be applied at inference time, using efficient MDM sampling for initial generation followed by \method{} self-correction as more tokens become visible. Because both modes share the same parameters, this approach would not require separate predictor and corrector networks. We leave the design and evaluation of such hybrid training and sampling strategies for future work.

\subsection{Finetuning from an MDM checkpoint}
\label{app:finetune}

Because the three streams of \method{} share their attention and MLP weights (Section~\ref{sec:method-efficient}), a pretrained MDM checkpoint can initialize all of them directly. We test whether a short finetune is enough to equip an existing MDM with self-correction. Starting from the MDLM checkpoint of Appendix~\ref{app:training-owt} after 200K steps on OWT, we either continue MDLM training for 50K steps, which gives the 250K-step MDLM used throughout Section~\ref{sec:sampling-finetuning}, or finetune the same checkpoint with the \method{} objective for 50K steps (sqrt noise schedule, learning rate $6 \times 10^{-4}$ with warmup and linear decay, global batch size 512, no beginning-of-sequence token). Both models see the same training data; the \method{} finetune costs $2.2\times$ the FLOPs per step, i.e., $1.25\times$ the total training FLOPs of the 250K-step MDLM.

\begin{figure*}[h]
    \centering
    \includegraphics[width=\textwidth]{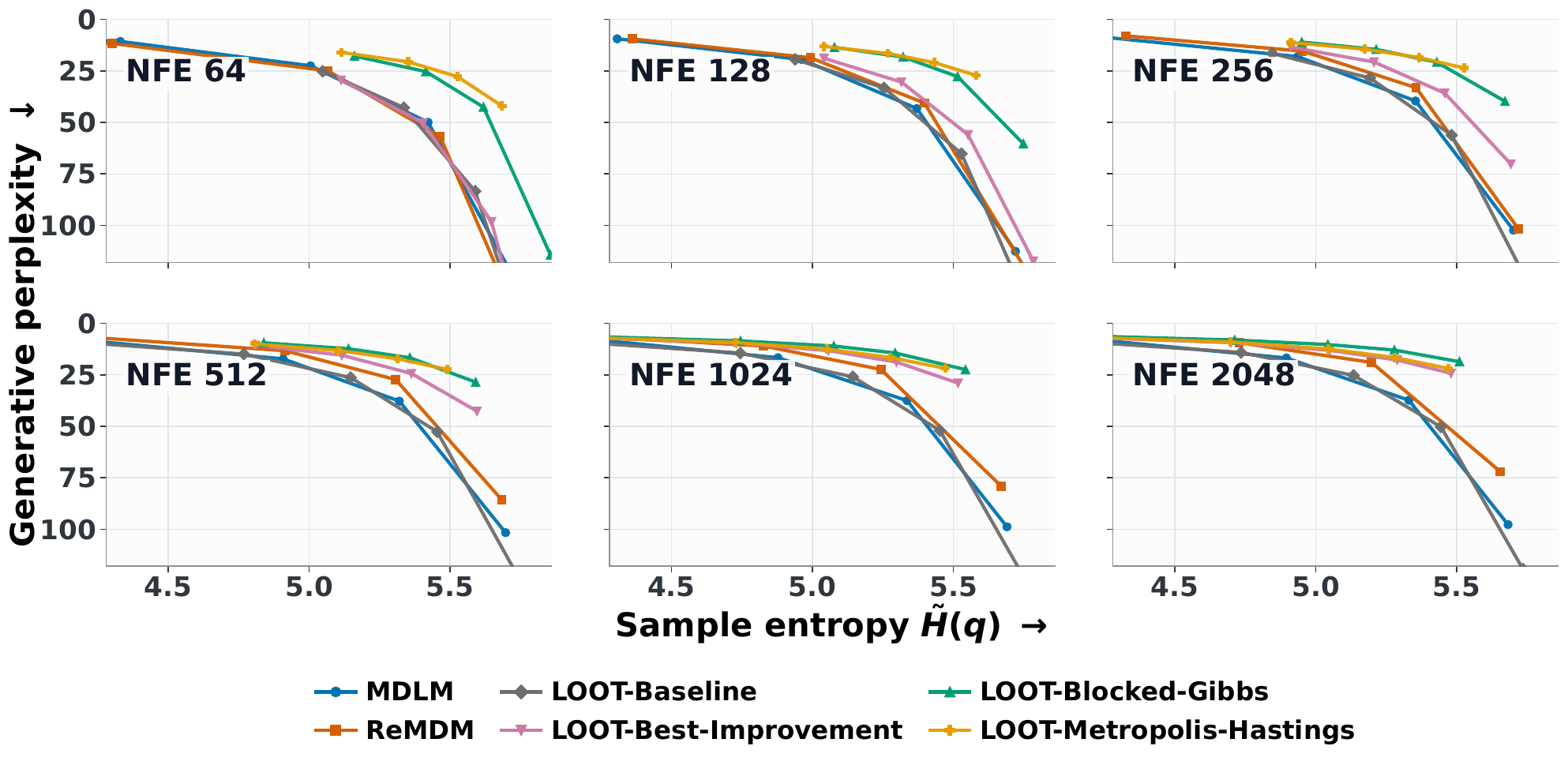}
    \caption{\textbf{Quality-diversity frontiers for \method{} finetuned from an MDM checkpoint.} Same protocol as Figure~\ref{fig:sampling-frontier-all-nfe} (top-$p = 0.9$); here the \method{} model is initialized from the 200K-step MDLM checkpoint and finetuned for 50K steps, while MDLM and ReMDM use the same checkpoint continued for 50K steps.}
    \label{fig:frontiers-finetune}
\end{figure*}

Figure~\ref{fig:frontiers-finetune} repeats the comparison of Figure~\ref{fig:sampling-frontier-all-nfe} with this finetuned model. Without correction, the finetuned \method{} tracks the continued-MDLM frontier at every budget, so 50K steps of leave-one-out training cost nothing in ancestral sampling quality. With correction, \method{} achieves a favorable generative perplexity--entropy tradeoff compared to ReMDM across all NFEs. An existing MDM checkpoint can thus be given self-correction with a finetune of a quarter of its pretraining steps.

\section{Computational cost}
\label{app:computational-cost}

\paragraph{FLOPs per forward pass.} Counting matmul and attention FLOPs at $N=1024$, a \method{} forward pass costs $2.24\times$ that of a DiT of the same size. The ratio is below the naive $3\times$ of three streams because \method{} doubles rather than triples the attention work (the forward and backward streams are batched into one causal attention call and the merge adds one sliding-window call per layer), uses a cheaper final merge block, and shares the vocabulary head. Since forward and backward passes scale together, training \method{} from scratch for the same number of steps as MDLM, as in Section~\ref{sec:sampling-finetuning}, costs $2.24\times$ the training FLOPs.

\paragraph{Measured throughput.} Table~\ref{tab:latency-b1} reports single-sequence latency without compilation. Table~\ref{tab:throughput-sweep} reports the forward+backward throughput ratio with \texttt{torch.compile} across batch sizes: at small batch both models are kernel-launch-bound and compilation fuses \method{}'s merge operations, so the overhead is $1.25$--$1.5\times$; at large batch and $N=1024$ both models are compute-bound and the ratio converges to $2.22\times$, matching the analytic FLOP ratio.

\begin{table}[h]
\centering
\caption{Latency of forward and backward passes on an NVIDIA B200 in bf16 at batch size 1, without compilation. We
report the median of 200 timed iterations.}
\label{tab:latency-b1}
\begin{tabular}{llrr}
\toprule
Dataset & Architecture & Fwd (ms) & Fwd+Bwd (ms)  \\
\midrule
\multirow{2}{*}{LM1B ($N{=}128$)}
  & MDM    & $13.3 \pm 0.4$ & $34.0 \pm 1.1$ \\
  & LOOT (ours)   & $28.1 \pm 0.2$ & $61.2 \pm 0.9$  \\
\midrule
\multirow{2}{*}{OWT ($N{=}1024$)}
  & MDM    & $13.4 \pm 0.1$ & $41.2 \pm 1.4$ \\
  & LOOT (ours)   & $22.1 \pm 0.1$ & $77.2 \pm 1.2$  \\
\bottomrule
\end{tabular}
\end{table}

At this operating point LOOT runs at $1.80\times$ MDM latency on LM1B ($N=128$) and $1.87\times$ on OWT ($N=1024$); Table~\ref{tab:throughput-sweep} shows that this single point understates the overhead at training batch sizes.

\begin{table}[h]
\centering
\caption{Forward+backward throughput ratio MDM/\method{} (higher means larger \method{} overhead) on an NVIDIA B200 with \texttt{torch.compile}, mean $\pm$ std over timed iterations.}
\label{tab:throughput-sweep}
{\footnotesize\setlength{\tabcolsep}{3pt}
\begin{tabular}{lccccccc}
\toprule
Dataset & $B{=}1$ & $B{=}2$ & $B{=}4$ & $B{=}8$ & $B{=}16$ & $B{=}32$ & $B{=}64$ \\
\midrule
LM1B ($N{=}128$) & $1.26 \pm 0.08$ & $1.29 \pm 0.09$ & $1.39 \pm 0.14$ & $1.25 \pm 0.11$ & $1.38 \pm 0.04$ & $1.38 \pm 0.13$ & $1.47 \pm 0.11$ \\
OWT ($N{=}1024$) & $1.27 \pm 0.08$ & $1.49 \pm 0.09$ & $1.40 \pm 0.12$ & $1.29 \pm 0.09$ & $2.18 \pm 0.01$ & $2.22 \pm 0.01$ & $2.22 \pm 0.01$ \\
\bottomrule
\end{tabular}}
\end{table}

\paragraph{Inference cost per sample.} Table~\ref{tab:inference-cost} reports counted FLOPs and measured wall-clock time per generated sequence at $N=1024$ on an NVIDIA A6000 (95\% intervals over 3 runs). All methods run at their common reliable batch size: MDLM and \method{} at batch 4, ReMDM at its memory ceiling of batch 3 (\method{} at batch 3 is essentially unchanged, at 11.52 and 10.48 seconds for the two correctors). \method{} with self-correction at 256 NFE uses $3.5\times$ fewer FLOPs ($71\%$) and about $8\times$ less wall-clock time per sample than ReMDM at 2048 NFE; at 512 NFE the FLOP saving is $1.7\times$ ($42\%$). Against MDLM at the same number of forward passes, corrected \method{} is end-to-end no slower despite $2.24\times$ the FLOPs per pass; the uncorrected row shows this is a property of the corrected samplers, not of the architecture alone. The correction budget is also compressible: 8 Metropolis correction steps already capture $85\%$ of the full 128-step improvement in generative perplexity.

\begin{table}[h]
\centering
\caption{Inference cost per generated sequence at $N=1024$ (NVIDIA A6000, bf16). FLOPs are counted matmul and attention FLOPs; seconds are measured end-to-end, with 95\% intervals over 3 runs.}
\label{tab:inference-cost}
\begin{tabular}{lccc}
\toprule
\textbf{Method} & \textbf{Steps (init + corr.)} & \textbf{TFLOPs / sample} & \textbf{Seconds / sample} \\
\midrule
MDLM & 256 + 0 & 75.0 & 11.76 [11.45, 12.08] \\
\method{}, no correction & 256 + 0 & 167.9 & 14.54 [14.38, 14.70] \\
\method{}-Blocked-Gibbs & 128 + 128 & 167.9 & 11.46 [11.15, 11.77] \\
\method{}-Metropolis-Hastings & 128 + 128 & 167.9 & 10.46 [10.17, 10.75] \\
ReMDM & 2048 + 0 & 583.3 & 95.54 [87.82, 103.26] \\
\bottomrule
\end{tabular}
\end{table}

\end{document}